\documentclass{article} %

\PassOptionsToPackage{table}{xcolor}
\usepackage{xcolor}
\usepackage{booktabs}
\usepackage{multirow}
\usepackage{makecell}
\usepackage{colortbl}
\usepackage{graphicx}
\usepackage{wrapfig}
\usepackage{pifont}
\definecolor{RedOrange}{rgb}{1,0.5,0}
\definecolor{BlueGreen}{rgb}{0.0,0.5,0.5}
\definecolor{CadetBlue}{rgb}{0.37,0.62,0.63}
\newcommand{\std}[1]{{\scriptsize\color{gray}$\pm$#1}}
\newcommand{\best}[1]{\textbf{#1}}
\newcommand{\second}[1]{\underline{#1}}
\newcommand{\model}[1]{\texttt{\small #1}}
\newcommand{\red}[1]{\textcolor{RedOrange}{$\uparrow$#1}}
\newcommand{\blue}[1]{\textcolor{BlueGreen}{$\downarrow$#1}}

\providecommand{\logo}[1]{\makebox[1.5em][l]{\raisebox{-0.3\height}{\includegraphics[height=2ex]{logos/#1}}}}

\usepackage{iclr2027_conference,times}

\usepackage{amsmath,amsfonts,bm}

\def\eqref#1{equation~\ref{#1}}

\def\1{\bm{1}}

\DeclareMathAlphabet{\mathsfit}{\encodingdefault}{\sfdefault}{m}{sl}
\SetMathAlphabet{\mathsfit}{bold}{\encodingdefault}{\sfdefault}{bx}{n}

\usepackage{hyperref}
\usepackage{url}
\usepackage{enumitem}
\usepackage{etoc}
\usepackage{placeins}
\usepackage[most]{tcolorbox}

\definecolor{codeteal}{HTML}{2B6F74}
\definecolor{coderose}{HTML}{9E4A5E}
\definecolor{codeslate}{HTML}{66788A}
\definecolor{codebg}{HTML}{F7F9FA}
\definecolor{codeframe}{HTML}{3F7F83}
\newcommand{\codefont}{\fontencoding{T1}\def\ttdefault{zi4}\ttfamily\fontsize{7.4}{9.2}\selectfont}
\lstdefinestyle{harness}{
  language=Python, basicstyle=\codefont,
  keywordstyle=\color{codeteal}\bfseries, morekeywords={None,True,False},
  stringstyle=\color{coderose}, commentstyle=\color{codeslate},
  showstringspaces=false, upquote=true, columns=fixed, basewidth=0.5em, keepspaces=true,
  breaklines=true, tabsize=4,
}
\newtcblisting{harnesscode}[2]{
  enhanced, breakable, listing only, listing options={style=harness},
  colback=codebg, colframe=codeframe, colbacktitle=codeframe, coltitle=white,
  boxrule=0.5pt, arc=1.2mm, left=2mm, right=2mm, top=0.8mm, bottom=0.8mm,
  fonttitle=\footnotesize\bfseries, title={#1\hfill{\mdseries\ttfamily #2}},
}

\title{Learning from Research:\\Toward Lifelong Agent Harness Evolution}

\author{Jingbo Yang$^{1}$\thanks{Correspondence to: Jingbo Yang\texttt{<jingbo@ucsb.edu>}. Work performed while interning at Microsoft.}\quad Kwei-Herng Lai$^{2}$\quad Xiaowen Wang$^{2}$\quad Yaar Harari$^{2}$ \\ \textbf{Evgeniy Gabrilovich$^{2}$\quad Shiyu Chang$^{1}$} \\
$^{1}$University of California, Santa Barbara \quad
$^{2}$Microsoft \\
}

\iclrfinalcopy %
\begin{document}

\maketitle
\etocdepthtag.toc{mainmatter}

\begin{abstract}
Language agents are expected to solve increasingly complex tasks, creating a growing need for continual improvement. One promising approach is to evolve the agent harness, the software that governs tool use, memory management, and task execution, while keeping the underlying language model fixed. Recent methods automate this process by using a meta coding agent to modify the harness based on execution feedback. However, relying on that agent's existing knowledge and observed failures can restrict exploration and make adaptation reactive. Inspired by how human experts learn from the research literature for new solutions, we introduce \textsc{ScholarEvolve}, a framework that automatically draws on state-of-the-art research to guide harness evolution. \textsc{ScholarEvolve} organizes the harness evolution directions into functional modules and uses topic modeling to identify distinct improvement strategies for each module. It implements these strategies and evaluates their combinations to improve
task performance.  Moreover, the framework is designed to incorporate new publications over time, allowing research advances to drive proactive lifelong evolution. Experiments demonstrate improvements on AppWorld and $\tau^2$-Bench. \textsc{ScholarEvolve} raises Qwen3.5-27B task goal completion from 49.6\% to 63.6\% on AppWorld Challenge, and raises GPT-5.4-mini pass$^1$ from 72.7\% to 81.9\% on $\tau^2$-Bench Telecom. Code is available at \url{https://github.com/UCSB-NLP-Chang/ScholarEvolve}.
\end{abstract}

\section{Introduction}
\label{sec:introduction}

Language model agents are increasingly able to solve problems that require extended interaction with digital environments, from debugging software repositories to coordinating workflows across applications~\citep{merrill2026terminal,wang2025openhands,trivedi2024appworld}. 
As their capabilities grow, so do the demands placed on them: users expect agents to handle longer tasks, unfamiliar tools, and changing requirements with greater autonomy~\citep{liu2026well}. Meeting these demands requires agents to acquire new problem solving strategies and improve how they use their available capabilities. This has motivated growing interest in \emph{agent evolution}, where an agent's design continues to improve beyond its initial development~\citep{yang2026federatedskill,zhang2026hyperagents,zhang2026darwin}.
Repeatedly training large backbone models can make such adaptation computationally expensive~\citep{zhang2026darwin,agrawal2026gepa}. Recent work therefore explores evolution of the \emph{agent harness}, the software that organizes tool use, context management, skills, memories, and execution workflows around a fixed model~\citep{team2026kimi,lee2026meta,lou2026autoharness}. These modules determine how the agent turns model outputs into sustained action, including what information it retains and how it recovers from failure~\citep{zhang2026agentic,ouyang2026reasoningbank,zhou2026memento}. Evolving the harness opens a practical path to more capable agents through changes to this execution system.

\begin{figure}[!t]
    \centering
    \includegraphics[width=\linewidth]{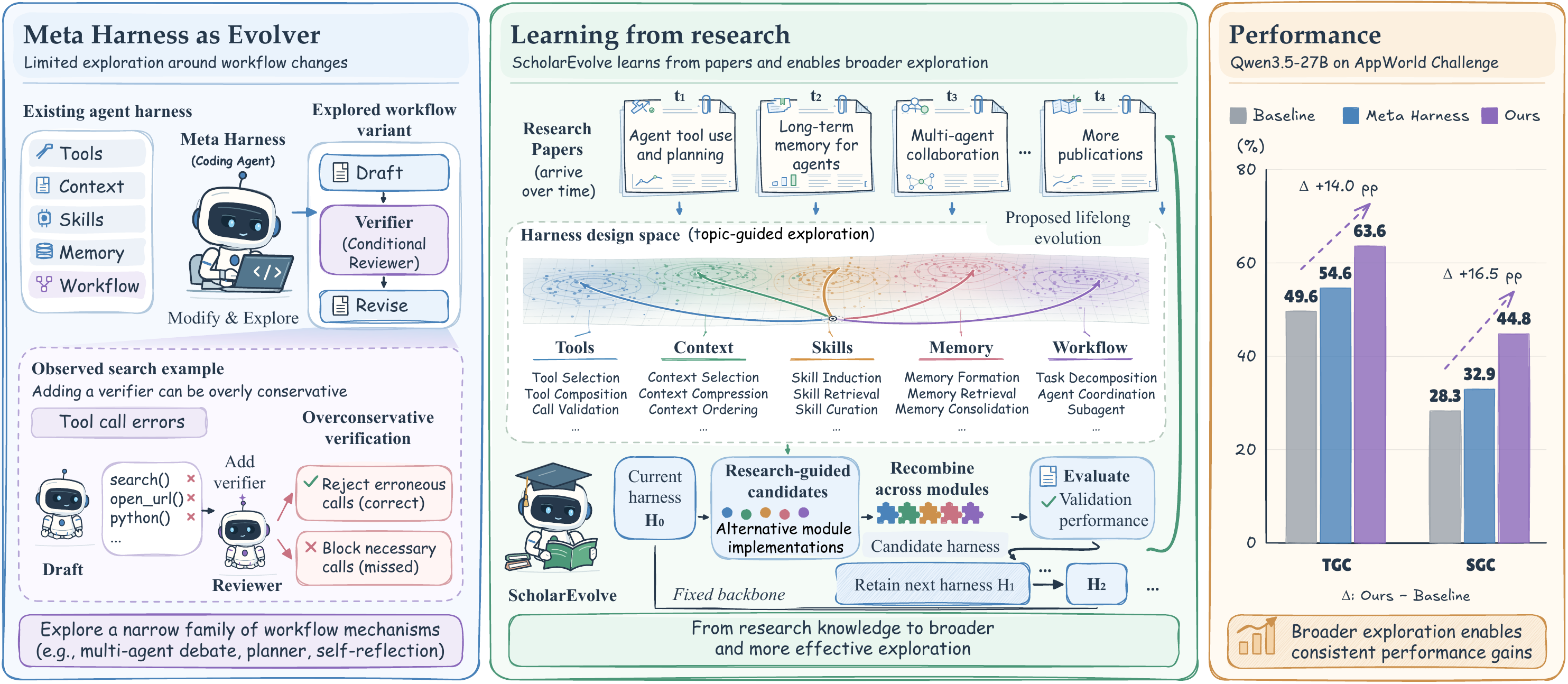}
    \vspace{-15pt}
    \caption{Overview of \textsc{ScholarEvolve}.
    Left: workflow focused exploration. Middle: research guided module
    mutation, recombination, and lifelong evolution.
    Right: Qwen3.5-27B task (TGC) and scenario (SGC) goal completion
    on AppWorld Challenge.}
    \label{fig:scholarevolve_intro}
\end{figure}

Recent harness evolution methods commonly use a \emph{meta agent}, often a coding agent equipped with its own development harness, to inspect execution traces, propose code changes, and evaluate the resulting candidates~\citep{lee2026meta,robeyns2025self}.
This approach automates an increasingly broad range of engineering decisions, yet sustaining progress raises three challenges. \ding{182}~\textbf{Limited exploration.} Broad permission to edit code can
still lead to a narrow sequence of modifications. Recent evaluations document
search that plateaus around repeated local edits~\citep{huang2026evo}. For example, adding verification
rounds or multiple agent proposals can increase computation while leaving
other mechanisms for improving the harness unexplored~\citep{cemri2026multi,kim2025towards}.
Figure~\ref{fig:scholarevolve_intro} (left) illustrates how overly conservative
verification can block tool calls needed to complete a task.
\ding{183}~\textbf{Limited design knowledge.} The mechanisms a meta agent
proposes depend on its backbone's knowledge and reasoning
capabilities~\citep{zhang2026darwin}. A failure trace may expose a missing capability while
offering little guidance on the technique needed to build it.
\ding{184}~\textbf{Delayed adaptation.} Evolution driven by trajectories and
feedback draws its evidence from situations already encountered~\citep{karten2026continual,wei2026evo}.
In deployment, a weakness may consequently affect users before sufficient
evidence accumulates to motivate a useful update. These challenges
point to the need for fresh design ideas beyond an agent's existing
experience. This leads us to ask: \textit{How would an experienced researcher
uncover improvements that an agent's own experience has yet to reveal?}

Experienced researchers and machine learning engineers turn to the research literature to learn how others solve similar problems, identify promising mechanisms, and adapt them to new settings~\citep{tang2026ai,schmidgall2025agentrxiv,gottweis2025towards}. This practice makes the collective experience of the research community available to each new design effort.
We bring this practice into \textsc{ScholarEvolve}, an automated research framework designed for lifelong agent harness evolution (Figure~\ref{fig:scholarevolve_intro}, middle), with three corresponding advantages. \ding{182}~\textbf{Structured exploration.} We organize search by harness module and use topic modeling to group papers into distinct mechanism families within each module. We merge overlapping topics and distribute the candidate budget across the remaining families to cover different mechanisms. Common interfaces support module recombination, and direct evaluation of the resulting combinations measures their joint effects before selection.
\ding{183}~\textbf{Proposals grounded in research.} Retrieved papers supply concrete techniques that guide the coding agent's implementation of module mutations. New publications can introduce mechanisms developed after the backbone was trained, expanding the design knowledge available to the meta agent without updating its model weights.
\ding{184}~\textbf{Proactive lifelong evolution.} The proposed lifelong extension uses a genetic algorithm in which research informs module mutations, recombination provides crossover, and validation selects the next champion. Periodic literature refresh can initiate this process as new methods become available. A deployed agent could, for example, review the latest research each month and evaluate promising updates against validation tasks.

We evaluate \textsc{ScholarEvolve}'s module search and composition on AppWorld~\citep{trivedi2024appworld} and $\tau^2$-Bench~\citep{barres2025tau}, using Qwen3.5-27B and GPT-5.4-mini as task agent backbones (Table~\ref{tab:main}). On AppWorld, the evolved Qwen3.5-27B harness raises task goal completion from 69.0\% to 81.4\% on the Normal split and from 49.6\% to 63.6\% on the Challenge split. On $\tau^2$-Bench Telecom, the evolved GPT-5.4-mini harness improves pass$^1$ from 72.7\% to 81.9\%.

\section{Related Work}
\label{sec:related_work}

\textbf{Agent harness evolution.}
Automated agent design searches for effective execution mechanisms around
language models. ADAS searches over agent programs, and AFlow optimizes
workflows through tree search~\citep{hu2025automated,zhang2025aflow}. Their evaluations emphasize static reasoning, question answering, and code generation, leaving sustained interaction with changing environment states less explored. Subsequent work broadens the optimization scope. DGM evolves coding agents through self-modification and an archive of candidates. Meta Harness searches executable harness programs using accumulated evaluation evidence, while AHE uses detailed trajectories to revise tools, memory, and other harness components~\citep{zhang2026darwin,lee2026meta,lin2026agentic}. AgentSquare also decomposes agents into modules and combines module evolution with recombination~\citep{shang2025agentsquare}. These approaches establish code search and modular composition as useful foundations. \textsc{ScholarEvolve} supplies this search with mechanisms extracted from the research literature, using topic modeling to organize distinct directions within each module and distribute exploration across them.

Generalization and continued adaptation are central challenges for evolved harnesses. Offline search typically yields a harness that is frozen for subsequent evaluation. Some reported gains also reuse the search tasks: Meta Harness's TerminalBench-2 experiment and AHE's primary evolution experiment optimize and report performance on the same 89 tasks~\citep{lee2026meta}. Such search can make the agent overfit to a group of specific tasks. Recent work addresses this distinction through regularized evolution in RRSI and separate development and final evaluation in SoL-Pi, which searches for composable efficiency improvements~\citep{liu2026sol}. For continued adaptation, Adaptive Auto-Harness and Evo-Harness update harnesses from experience accumulated over task streams~\citep{liu2026adaptive,wei2026evo}. \textsc{ScholarEvolve} introduces a complementary source of updates: new publications can initiate further generations of module mutation and recombination, allowing a deployed agent to evaluate new mechanisms before corresponding failures accumulate in its own interactions.

\textbf{Automated research.}
Automated research systems turn scientific knowledge and experimentation into reusable discoveries. AI Scientist-v2 and AI-Researcher automate hypothesis generation, implementation, experimentation, and manuscript preparation~\citep{yamada2025ai,tang2026ai}. AgentRxiv enables research agents to share reports and build on earlier findings, demonstrating cumulative improvements in reasoning and prompting methods~\citep{schmidgall2025agentrxiv}. Paper2Agent converts papers and associated code into callable tools and interactive research agents, making published methods directly reusable~\citep{miao2026reimagining}. \textsc{ScholarEvolve} makes the translation from literature to harness improvements an explicit search process. Papers provide candidate mechanisms, topic modeling organizes their coverage, and module interfaces support implementation and recombination. Task evaluation selects useful changes, while successive literature updates provide fresh directions for lifelong evolution around a fixed backbone.

\section{Method}
\label{sec:method}

\begin{figure}[!t]
    \centering
    \includegraphics[width=\linewidth]{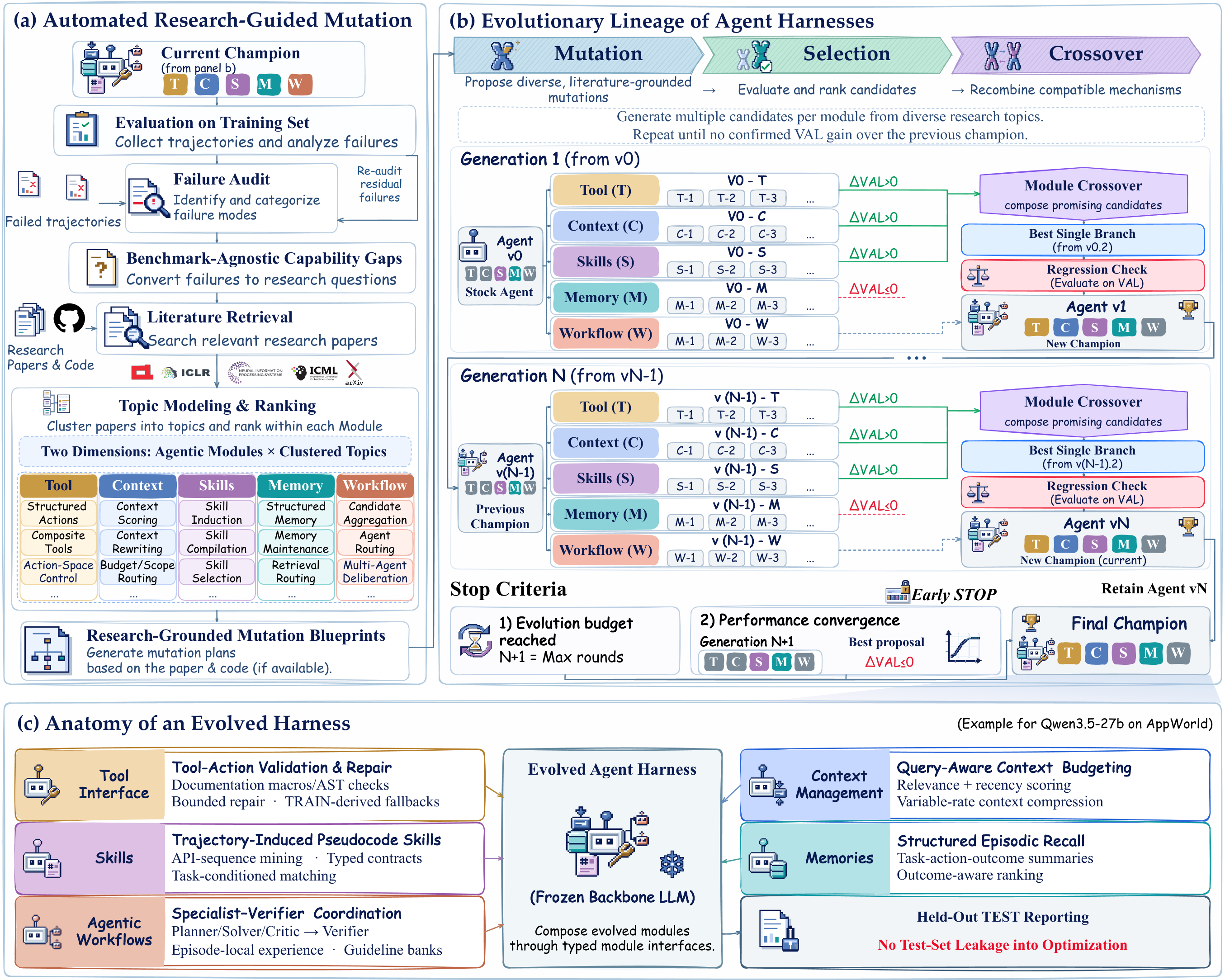}
    \vspace{-15pt}
    \caption{\textsc{ScholarEvolve}: (a) research guided module mutation,
    (b) selection and crossover across generations, and (c) evolved modules
    operating around a fixed backbone.}
    \label{fig:ripe_method}
\end{figure}

We propose \textsc{ScholarEvolve}, a framework for lifelong harness evolution
through literature guided module mutation and crossover around a fixed
backbone (Figure~\ref{fig:ripe_method}).

\subsection{Problem formulation}
\label{sec:formulation}

An \emph{agent harness} is an executable program that manages model inputs,
action execution, and state updates during interaction with an
environment~\citep{ning2026code}. Let $\mathcal H$ contain harnesses that
satisfy the environment's interfaces and execution constraints.

We distinguish the target task distribution from the finite splits available
to evolution. For each evaluation setting, let $\mathcal P_{\mathrm{tar}}$
denote the task distribution on which performance is ultimately desired. The
\emph{evolution set} $\mathcal D_{\mathrm{evo}}$ supplies trajectories for
failure analysis, research query construction, and persistent artifact
construction. The \emph{validation set} $\mathcal D_{\mathrm{val}}$ supports
candidate evaluation and champion selection. The \emph{test set}
$\mathcal D_{\mathrm{test}}$ is a finite held-out sample from
$\mathcal P_{\mathrm{tar}}$, used only for final
reporting after all selection decisions are frozen. Thus,
$\mathcal D_{\mathrm{test}}$ represents the target distribution in evaluation,
but is not the distribution itself. The three sets are mutually disjoint.

The \emph{fixed backbone} $f_\theta$ is the language model whose parameters
remain unchanged during evolution. For a task $x\sim\mathcal P_{\mathrm{tar}}$,
harness $H$ induces an interaction trajectory
$\tau=(o_0,a_0,\ldots,o_T)$ containing observations, model decisions, tool
calls, and environment responses. We write
$\tau\sim p_\theta(\cdot\mid H,x)$ and score it by reward $R(x,\tau)$.
Harness evolution seeks
\begin{equation}
    H^\star\in\operatorname*{arg\,max}_{H\in\mathcal H}J(H),\qquad
    J(H)=\mathbb E_{x\sim\mathcal P_{\mathrm{tar}}}\,
    \mathbb E_{\tau\sim p_\theta(\cdot\mid H,x)}[R(x,\tau)].
    \label{eq:harness}
\end{equation}
Starting from a base harness $H^{(0)}$, the search uses
$\mathcal D_{\mathrm{evo}}$ and $\mathcal D_{\mathrm{val}}$ within a fixed
budget. It never uses $\mathcal D_{\mathrm{test}}$ or updates $\theta$.

\subsection{Research guided modular exploration}
\label{sec:module_mutation}

We organize exploration along two axes: a \emph{module} specifies the
execution responsibility to modify, and a \emph{method topic} identifies
a research mechanism to implement within it. Together they distribute
candidate generation across intervention points and alternative solutions.

\subsubsection{Structuring the code space into modules}
\label{sec:modules}

We represent a harness as $H=\operatorname{Compose}(m_1,\ldots,m_5)$,
where each $m_j$ implements a defined interface. The five modules provide
separate intervention points and exchange boundaries for crossover.
Their definitions follow the modular harness view in modern agent
post-training~\citep{team2026kimi}.

\ding{182}~\textbf{Tool interface.} Controls the actions available to the
agent and translates proposals into executable calls. Mutation directions
include \emph{tool selection}, \emph{tool composition}, and \emph{call
validation}, such as exposing task relevant tools or packaging related
operations into a composite action.
\ding{183}~\textbf{Context management.} Assembles model inputs from
instructions, history, memories, and skills through \emph{context selection},
\emph{compression}, and \emph{ordering}, preserving evidence needed for
subsequent decisions within the available context budget.
\ding{184}~\textbf{Skills.} Maintains reusable procedures through \emph{skill
induction}, \emph{retrieval}, and \emph{curation}. Mutations can extract
procedures from successful trajectories, retrieve them by applicability,
and consolidate overlapping entries in a skill library.
\ding{185}~\textbf{Memories.} Maintains experience through \emph{memory
formation}, \emph{retrieval}, and \emph{consolidation}, for example by
recording episode outcomes, recalling relevant situations, and merging
related records.
\ding{186}~\textbf{Agentic workflows.} Organizes planning and action
generation through \emph{task decomposition}, \emph{agent coordination},
and \emph{subagent delegation}, including assigning subtasks and integrating
specialist proposals.

These responsibilities interact during execution: retrieving a longer skill
changes what context management must preserve, and delegating a subtask
changes the evidence available to subsequent decisions. During mutation,
we change one module while keeping the others fixed, isolating its effect
within the current harness. Coordination among independently modified
modules is handled during crossover, where complete harnesses are evaluated
to select compatible combinations (Section~\ref{sec:module_composition}).

\subsubsection{Constructing the research pool}
\label{sec:module_research}

As shown in Figure~\ref{fig:ripe_method}(a), research pool construction begins
with trajectories collected on $\mathcal D_{\mathrm{evo}}$. An auditor
attributes failed trajectories to agent or environment causes and retains recurring
agent-attributable failures together with their supporting observations and
actions. A research model abstracts this benchmark-specific evidence into
capability gaps that describe missing inference-time abilities. For each gap,
it identifies possible interventions within each module's responsibility and
generates broad capability queries and focused mechanism queries. Literature
retrieval and deduplication then produce a pool $\mathcal P_j$ of paper titles
and abstracts for module $j$, ready for mechanism level topic modeling.
Appendix~\ref{app:research_pipeline} details the audit, query construction,
retrieval, and screening.

\subsubsection{Topic modeling for orthogonal exploration}
\label{sec:module_topics}

A broad research pool can contain many papers proposing similar mechanisms.
Spending the mutation budget on these papers narrows exploration, while
overlapping interventions can complicate coordination during crossover.
To broaden exploration and mitigate such conflicts, we organize papers into
semantically orthogonal topics, each describing a distinct mechanism with
minimal overlap in its intended operation. We follow the TopicGPT
framework~\citep{pham2024topicgpt} for prompted topic generation, refinement,
and assignment, conditioning each stage on the module's responsibility
and encouraging consistent mechanism granularity. 

\textbf{Taxonomy induction and refinement.}
Let $d_j$ describe module $j$, and $B_{j,b}$ be batch $b$ of its papers.
Starting from $\mathcal T_j^{(0)}=\emptyset$, the research model induces
and refines topics through
\begin{equation}
    \mathcal T_j^{(b)}=\mathcal T_j^{(b-1)}\cup
      G_\phi\!\left(B_{j,b},\mathcal T_j^{(b-1)},d_j\right),
    \qquad
    \mathcal T_j=F_\phi\!\left(\mathcal T_j^{(B_j)},d_j\right),
    \label{eq:topic_induction}
\end{equation}
where $B_j$ is the number of batches. Each topic has a mechanism name and
a short description. Generation $G_\phi$ reuses a name when its mechanism
fits a paper and adds a category when a relevant mechanism remains
uncovered. Refinement $F_\phi$ merges paraphrases and overlapping categories,
absorbs overly specific variants into broader mechanisms, and removes
vague or irrelevant topics. For example, demonstration retrieval and
knowledge retrieval can be grouped under \emph{External Context Retrieval}
when retrieval is the shared intervention. This standardization of
mechanism granularity encourages semantic orthogonality across topics.

\textbf{Evidence supported assignment.}
The research model assigns each paper an ordered list
$L_j(p)=(\ell_{j,p,1},\ldots,\ell_{j,p,q_{j,p}})$ of topics, placing its
primary methodological contribution first and providing supporting
excerpts from the abstract. We retain labels in the refined taxonomy.
For a nonempty list, the primary topic $z_j(p)=\ell_{j,p,1}$ defines the
selection cluster $\mathcal C_{j,t}=\{p\in\mathcal P_j:z_j(p)=t\}$.
Papers without a retained label remain unassigned. Multiple labels
preserve a paper's different mechanisms, while the primary label gives
it one selection cluster per module. Stored excerpts make these
assignments inspectable.

\textbf{Coverage across mechanism families.}
We screen topics against mechanisms already available in the base harness
and backbone. Let $\mathcal T_j^+$ contain the remaining topics with
nonempty clusters. We order clusters by decreasing size and select papers
in round robin order, visiting every cluster before returning to one.
Within a cluster, we follow retrieval order. A budget of $K_j$ papers
therefore covers $\min(K_j,|\mathcal T_j^+|)$ distinct topics.
Topic refinement reduces redundant mechanisms, while this allocation
spreads implementation opportunities across them. Topic orthogonality
and module isolation support crossover at complementary levels: topics
diversify the implementations available within a module, and interfaces
localize their responsibilities. For example, context compression and
workflow verification can be composed through separate interfaces.
Their joint effectiveness is assessed by evaluating the complete harness.

\subsubsection{From research mechanisms to executable mutations}

For each selected paper $p$, a research agent reads the full paper,
including its method and appendix, together with available official code.
It produces a \emph{mutation blueprint} that specifies the source mechanism
and assumptions, target module and interface, required state and artifacts,
runtime operations, auxiliary model calls, and termination conditions.
The blueprint separates the source mechanism from adaptations required by
the host harness and retains the paper and topic provenance.

The coding agent receives this blueprint, the current module implementation,
its interface, and the environment's action format. It implements
$m_j^{(p)}$ and constructs the mutation
$H_j^{(p)}=H^{(0)}[j\leftarrow m_j^{(p)}]$, keeping the other modules fixed.
Literature-derived mechanisms must be adapted to the host harness, so a
candidate can fail at an interface or artifact boundary before its behavior
can be measured. We therefore run lightweight health probes that check
importability, interface compatibility, artifact construction and reloading,
and valid action production. The coding agent repairs reported errors until
these checks pass or its repair budget is exhausted. Persistent skill libraries
and memories may be built only from trajectories in
$\mathcal D_{\mathrm{evo}}$. They are frozen during evaluation on
$\mathcal D_{\mathrm{val}}$ and $\mathcal D_{\mathrm{test}}$, while modules
can update local state within an episode.
Appendix~\ref{app:method_details} provides concrete implementations and
additional implementation details.

\subsection{Module crossover and selection}
\label{sec:module_composition}

At generation $g$, the retained \emph{champion} $H_g$ serves as the base
harness $H^{(0)}$. We first evaluate each mutation against it on the same
$\mathcal D_{\mathrm{val}}$ tasks and trials. With $\bar r_i(H)$ denoting mean
reward on task $i$, its gain is
$g_{j,p}(i)=\bar r_i(H_j^{(p)})-\bar r_i(H^{(0)})$.
\emph{Crossover} chooses one implementation per module. A configuration
$c=(c_1,\ldots,c_5)$ may retain a base module by setting $c_j=0$ and
$g_{j,0}(i)=0$. We rank combinations by additive predicted gain on
$\mathcal D_{\mathrm{val}}$:
\begin{equation}
    \widehat\Delta(c)=\frac{1}{|\mathcal D_{\mathrm{val}}|}
       \sum_{i\in\mathcal D_{\mathrm{val}}}\sum_{j=1}^{5}g_{j,c_j}(i).
    \label{eq:composition_prediction}
\end{equation}
We enumerate available combinations and use this score to shortlist them
within the crossover evaluation budget, reusing individual measurements
without additional rollouts. Each shortlisted configuration is assembled
as a complete harness $H_c$ and evaluated to obtain its actual gain
$\Delta(c)=|\mathcal D_{\mathrm{val}}|^{-1}
\sum_{i\in\mathcal D_{\mathrm{val}}}
[\bar r_i(H_c)-\bar r_i(H^{(0)})]$.
This joint evaluation provides the regression check shown in
Figure~\ref{fig:ripe_method}: selection uses observed gains to account for
module interactions and reward ceilings. Retaining the base implementation
at each position permits subsets of mutations, including omission of a
module whose individual gain fails to carry over to the combination.

Let $\mathcal A_g$ contain the evaluated mutations and crossover candidates, and
let $\widetilde H_g$ maximize their mean validation reward. We compare it
with $H_g$ using paired task gains. If $L_\alpha(H,H_g)$ is the lower
endpoint of their paired bootstrap gain interval at nominal confidence
level $1-\alpha$, the retention rule is
\begin{equation}
    H_{g+1}=\begin{cases}
        \widetilde H_g,&L_\alpha(\widetilde H_g,H_g)>0,\\
        H_g,&\text{otherwise}.
    \end{cases}
    \label{eq:champion_update}
\end{equation}
The champion is retained when no executable candidate is available.
Including individual mutations allows advancement through a single useful
change, corresponding to the best single branch in
Figure~\ref{fig:ripe_method}. The selected code and its frozen artifacts
form the next champion. We declare the current search cycle converged
after $s_{\mathrm{stop}}$ consecutive generations without a confirmed
validation improvement under Equation~\ref{eq:champion_update}. The cycle
also stops when its search budget is exhausted.

\subsection{Lifelong harness evolution}
\label{sec:champion_evolution}

Literature based evolution naturally supports \emph{proactive adaptation}:
new publications provide candidate improvements before corresponding
failures are observed in a deployed agent. Periodic literature updates
can therefore initiate a new search cycle even after the previous cycle
has converged. Each cycle starts from the retained champion and screens
research topics against its current mechanisms. Available trajectories from
$\mathcal D_{\mathrm{evo}}$ can refine the research brief as the harness evolves.

Within each cycle, literature supplies mutation directions, crossover
combines module implementations, and $\mathcal D_{\mathrm{val}}$ provides the
fitness signal for selection. The backbone remains fixed across cycles, while
the champion's code and persistent artifacts carry forward useful
improvements.

\section{Experiments}
\label{sec:experiments}

\subsection{Experimental Setup}
\label{sec:experimental_setup}

\textbf{Benchmarks and environments.}
AppWorld~\citep{trivedi2024appworld} evaluates tasks across applications through API execution.
Its 90 training tasks form $\mathcal D_{\mathrm{evo}}$, and its 57 development
tasks form $\mathcal D_{\mathrm{val}}$. We report separately on the official
Normal and Challenge test sets, denoted
$\mathcal D_{\mathrm{test}}^{\mathrm{N}}$ (168 tasks) and
$\mathcal D_{\mathrm{test}}^{\mathrm{C}}$ (417 tasks). Challenge requires APIs
from Amazon or Gmail, applications absent from the evolution and validation
sets.
We report task goal completion (TGC) and scenario goal completion (SGC),
which requires success on all three variants of a scenario.
For $\tau^2$-Bench~\citep{barres2025tau}, we use Telecom, where agents resolve service
issues through tools and a simulated user. A stratified set of 250 training
tasks forms $\mathcal D_{\mathrm{evo}}$, the official 74-task training split
serves as $\mathcal D_{\mathrm{val}}$, and the official 40-task test split is
$\mathcal D_{\mathrm{test}}$. We report pass$^k$, the probability of success
across all $k$ trials, for $k\in\{1,2,3,4\}$.

\textbf{Implementation details.}
We evolve Qwen3.5-27B and GPT-5.4-mini harnesses with fixed task backbones.
Comparisons include the initial harness and Meta Harness~\citep{lee2026meta}, a coding
agent that revises the harness using execution feedback. Both evolution
methods use GPT-5.4 through Codex and receive the same allocated search
rollout budget per setting. We report means and standard deviations over
three independent evaluation runs. Appendix~\ref{app:implementation_details}
specifies inference settings, search configurations, and metric aggregation.

\subsection{Main Results}
\label{sec:main_results}

Table~\ref{tab:main} compares \textsc{ScholarEvolve}, Meta Harness, and the
initial harness under each backbone.

\providecommand{\std}[1]{{\scriptsize\color{gray}$\pm$#1}}      %
\providecommand{\best}[1]{\textbf{#1}}                            %
\providecommand{\second}[1]{\underline{#1}}                       %
\providecommand{\model}[1]{\texttt{\small #1}}                    %
\providecommand{\red}[1]{\textcolor{RedOrange}{$\uparrow$#1}}     %
\providecommand{\blue}[1]{\textcolor{BlueGreen}{$\downarrow$#1}}  %
\providecommand{\zero}[1]{\textcolor{gray}{#1}}                   %
\providecommand{\logo}[1]{\makebox[1.5em][l]{\raisebox{-0.3\height}{\includegraphics[height=2ex]{logos/#1}}}}  %

\begin{table}[t]
  \centering
  \small
  \setlength{\tabcolsep}{4.0pt}
  \renewcommand{\arraystretch}{1.15}
  \caption{Main results on \textbf{AppWorld} and \textbf{$\tau^2$-Bench} as
  mean $\pm$ standard deviation over 3 runs. $\Delta$ is the absolute change
  from the matched backbone (\textcolor{RedOrange}{orange}: improvement,
  \textcolor{BlueGreen}{blue}: degradation). \textbf{Bold} and \underline{underline}
  mark the two largest gains per column.}
  \label{tab:main}

  \resizebox{\linewidth}{!}{
  \begin{tabular}{cl|cc|cc|cccc}
    \toprule
    \rowcolor{CadetBlue!15}
    \cellcolor{white} & \cellcolor{white}
    & \multicolumn{4}{c|}{\textbf{AppWorld}}
    & \multicolumn{4}{c}{\textbf{$\tau^2$-Bench}} \\
    \cmidrule(lr){3-6} \cmidrule(lr){7-10}
    \cellcolor{white} & \cellcolor{white}
    & \multicolumn{2}{c|}{\textit{Normal}} & \multicolumn{2}{c|}{\textit{Challenge}}
    & \multicolumn{4}{c}{\textit{Pass$^k$}} \\
    \cmidrule(lr){3-4} \cmidrule(lr){5-6} \cmidrule(lr){7-10}
    \cellcolor{white} & \cellcolor{white} \multirow{-3.25}{*}{\textbf{Method}}
    & TGC & SGC & TGC & SGC
    & $k$=1 & $k$=2 & $k$=3 & $k$=4 \\
    \midrule

    \multirow{10}{*}{\makecell{\textsc{Baseline}\\\textsc{Agents}}}
    & \logo{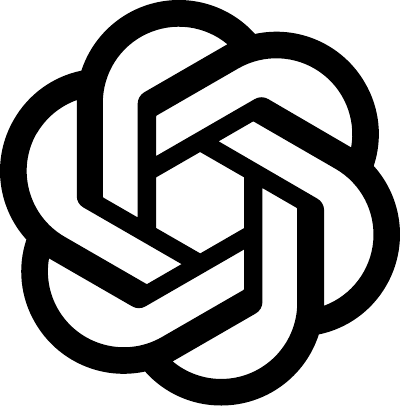}\model{gpt-5.4} & 85.7\std{2.73} & 73.2\std{5.46} & 80.4\std{1.05} & 63.3\std{3.31} & 86.3\std{2.89} & 75.6\std{6.38} & 67.2\std{10.29} & 60.0\std{14.43} \\
    & \logo{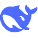}\model{DeepSeek-V4-Flash} & 84.1\std{2.27} & 68.4\std{7.22} & 79.5\std{0.51} & 60.4\std{0.75} & 98.1\std{0.00} & 96.3\std{0.00} & 94.4\std{0.00} & 92.5\std{0.00} \\
    & \logo{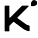}\model{Kimi-K2.6} & 81.3\std{0.91} & 60.1\std{2.73} & 77.5\std{1.04} & 59.0\std{3.14} &  98.1\std{0.88} & 97.9\std{0.59} & 97.7\std{0.30} & 97.5\std{0.00}  \\
    & \logo{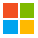}\model{MAI-Thinking-1} & 69.0\std{2.56} & 46.4\std{4.71} & 53.5\std{0.64} & 29.7\std{1.76} & 55.9\std{3.98} & 41.5\std{5.60} & 32.2\std{6.63} & 25.0\std{7.07} \\
    & \logo{openai}\model{gpt-oss-120b} & 34.5\std{1.20} & 14.3\std{3.12} & 21.6\std{0.29} & \phantom{0}7.4\std{2.08} & 74.4\std{0.88} & 60.2\std{0.88} & 50.3\std{0.44} & 42.5\std{0.00} \\
    & \logo{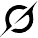}\model{grok-4.6} & 20.8\std{2.12} & \phantom{0}4.8\std{1.04} & 13.7\std{3.86} & \phantom{0}3.1\std{0.40} & 82.5\std{0.00} & 69.2\std{0.00} & 59.4\std{0.00} & 52.5\std{0.00} \\
    & \logo{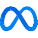}\model{Llama-3.3-70B} & 17.9\std{3.59} & \phantom{0}3.6\std{0.00} & \phantom{0}8.3\std{1.30} & \phantom{0}1.4\std{0.75} & 15.0\std{0.88} & 10.8\std{0.59} & 8.75\std{0.00} & 7.5\std{0.00} \\
    & \logo{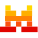}\model{Mistral-Large-3} & 17.3\std{1.15} & \phantom{0}7.0\std{1.66} & \phantom{0}7.4\std{0.87} & \phantom{0}0.9\std{0.40} & 35.4\std{5.67} & 23.2\std{6.38} & 15.6\std{7.21} & 10.8\std{7.64} \\
    & \logo{openai}\model{gpt-5.4-mini} & 67.1\std{1.78} & 46.4\std{3.06} & 46.1\std{0.98} & 21.1\std{1.11} & 72.7\std{2.53} & 61.1\std{4.26} & 54.4\std{6.03} & 49.2\std{8.04} \\
    & \logo{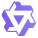}\model{Qwen3.5-27b} & 69.0\std{2.35} & 48.8\std{2.75} & 49.6\std{2.89} & 28.3\std{4.37} & 96.7\std{0.95} & 93.3\std{1.91} & 90.0\std{2.86} & 86.7\std{3.82} \\
    \midrule

    \multirow{4}{*}{\makecell{\textsc{Meta}\\\textsc{Harness}}}
    & Codex + \model{gpt-5.4-mini} & 67.5\std{2.44} & 45.8\std{2.70} & 45.6\std{0.42} & 17.8\std{0.83} & 66.5\std{0.95} & 51.7\std{2.53} & 44.0\std{3.08} & 39.2\std{2.89} \\
    & \quad $\Delta$ \textit{vs.}\ \model{gpt-5.4-mini} & \red{0.4} & \blue{0.6} & \blue{0.5} & \blue{3.3} & \blue{6.2} & \blue{9.4} & \blue{10.4} & \blue{10.0} \\
    & Codex + \model{Qwen3.5-27b}  & 76.8\std{1.19} & 62.5\std{3.57} & 54.6\std{0.97} & 32.9\std{1.50} & 96.7\std{1.30} & 93.6\std{2.37} & 90.8\std{3.21} & 88.3\std{3.82} \\
    & \quad $\Delta$ \textit{vs.}\ \model{Qwen3.5-27b} & \second{\red{7.8}} & \second{\red{13.7}} & \red{5.0} & \red{4.6} & \zero{0.0} & \red{0.3} & \red{0.8} & \red{1.6} \\
    \midrule

    \rowcolor{gray!10}
    \cellcolor{white}
    & \textsc{Ours} + \model{gpt-5.4-mini} & 72.4\std{1.83} & 55.4\std{3.55} & 55.6\std{2.10} & 32.4\std{2.15} & 81.9\std{1.65} & 71.5\std{2.29} & 64.2\std{3.15} & 58.3\std{3.82} \\
    \rowcolor{gray!10}
    \cellcolor{white}
    & \quad $\Delta$ \textit{vs.}\ \model{gpt-5.4-mini} & \red{5.3} & \red{9.0} & \second{\red{9.5}} & \second{\red{11.3}} & \best{\red{9.2}} & \best{\red{10.4}} & \best{\red{9.8}} & \best{\red{9.1}} \\
    \rowcolor{gray!10}
    \cellcolor{white}
    & \textsc{Ours} + \model{Qwen3.5-27b}  & 81.4\std{1.19} & 69.0\std{4.10} & 63.6\std{0.36} & 44.8\std{0.40} & 98.1\std{1.25} & 96.4\std{2.29} & 94.8\std{3.15} & 93.3\std{3.82} \\
    \rowcolor{gray!10}
    \cellcolor{white} \multirow{-4}{*}{\textsc{Ours}}
    & \quad $\Delta$ \textit{vs.}\ \model{Qwen3.5-27b} & \best{\red{12.4}} & \best{\red{20.2}} & \best{\red{14.0}} & \best{\red{16.5}} & \second{\red{1.4}} & \second{\red{3.1}} & \second{\red{4.8}} & \second{\red{6.6}} \\
    \bottomrule
  \end{tabular}
  }
\end{table}

\textbf{Consistent improvements across environments.}
\textsc{ScholarEvolve} improves both task backbones on every reported
AppWorld and Telecom metric. On AppWorld Normal, Qwen gains 12.4 TGC
and 20.2 SGC percentage points over the initial harness. On Telecom,
Mini gains 9.2 pass$^1$ and 9.1 pass$^4$ points. These improvements span
individual task completion and consistent success across variants or
repeated trials.

\textbf{Generalization to held-out applications.}
The same AppWorld harnesses and evolution-derived artifacts transfer to
Challenge, including the unseen applications Amazon and Gmail.
TGC gains are larger on Challenge than Normal: 14.0 versus 12.4 points
for Qwen, and 9.5 versus 5.3 for Mini. On Challenge, they exceed Meta
Harness by 9.0 and 10.0 points, respectively. Section~\ref{sec:evolution_analysis}
examines this transfer across API requirements relative to training.

\textbf{Closing capability gaps through harness design.}
On AppWorld Normal, the evolved Qwen agent reaches 81.4\% TGC, comparable
to Kimi-K2.6's 81.3\%, and reduces its gap to GPT-5.4 from 16.7 to 4.3
points, closing approximately 74\% of the gap with fixed model weights.
On Telecom, its 98.1\% pass$^1$ and 93.3\% pass$^4$ match or exceed
DeepSeek-V4-Flash's 98.1\% and 92.5\% under the initial harness. Harness design thus has a
substantial effect on the capabilities delivered by a given backbone.

\subsection{Lifelong Evolution}
\label{sec:lifelong_experiments}

\begin{figure}[t]
  \centering
  \includegraphics[width=\linewidth]{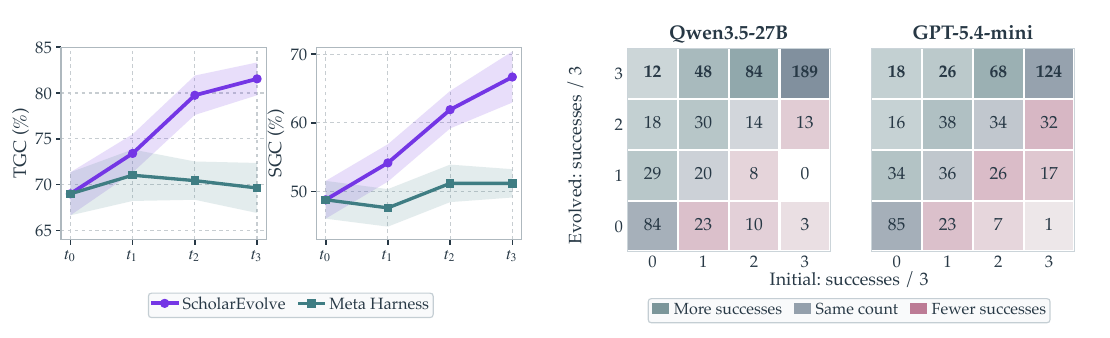}
  \vspace{-20pt}
  \caption{\textbf{Evolution over time and across held-out tasks.}
  Left: best harness found in each generation on AppWorld Normal, with bands
  showing one standard deviation. Right: task-level success-count transitions pooled
  over Normal and Challenge ($n=585$ per backbone).}
  \label{fig:evolution_summary}
\end{figure}
\begin{wraptable}{r}{0.50\textwidth}
  \vspace{-8pt}
  \centering
  \caption{Cumulative ablations on AppWorld Normal (\%). Each row
  additionally removes the named component. Scores are mean $\pm$ SD.}
  \label{tab:ablation}
  \vspace{3pt}
  \fontsize{8}{9.5}\selectfont
  \setlength{\tabcolsep}{2.5pt}
  \renewcommand{\arraystretch}{1.16}
  \newcommand{\abstd}[1]{{\fontsize{6}{7}\selectfont\color{gray}$\pm$#1}}
  \resizebox{\linewidth}{!}{%
  \begin{tabular}{lcccc}
    \toprule
    & \multicolumn{2}{c}{\textbf{Qwen3.5-27B}}
      & \multicolumn{2}{c}{\textbf{GPT-5.4-mini}} \\
    \cmidrule(lr){2-3}\cmidrule(lr){4-5}
    Variant & TGC $\uparrow$ & SGC $\uparrow$ & TGC $\uparrow$ & SGC $\uparrow$ \\
    \midrule
    \rowcolor{gray!10}
    \textsc{ScholarEvolve} & 81.4\abstd{1.19} & 69.0\abstd{4.10} & 72.4\abstd{1.83} & 55.4\abstd{3.55} \\
    $-$ topic-guided selection & 80.0\abstd{0.91} & 66.7\abstd{4.12} & 69.8\abstd{0.91} &
    50.0\abstd{4.72}\\
    \hspace{0.4em}$-$ module-wise mutation& 78.6\abstd{1.79} & 64.9\abstd{4.49} & 69.3\abstd{1.24} & 48.8\abstd{3.72} \\
    \hspace{0.8em}$-$ research guidance & 76.8\abstd{1.19} & 62.5\abstd{3.57} & 67.5\abstd{2.44} & 45.8\abstd{2.70} \\
    \bottomrule
  \end{tabular}%
  }
\vspace{-12pt}
\end{wraptable}

We introduce new papers over three publication windows while keeping
Qwen3.5-27B, its $\mathcal D_{\mathrm{evo}}$ trajectories, and its derived
artifacts fixed. 
Each round explores all five modules from the current harness.
Figure~\ref{fig:evolution_summary} (left) shows the initial harness and each
generation's best harness, selected on $\mathcal D_{\mathrm{val}}$ and evaluated on
$\mathcal D_{\mathrm{test}}^{\mathrm{N}}$. Both TGC and SGC improve across
all three updates, reaching 81.5\% and 66.7\%, respectively. 
The final
advantages over Meta Harness are 11.9 and 15.5 points.
Appendix~\ref{app:lifelong} provides the protocol and complete results.

\subsection{Ablation Studies}
\label{sec:ablation_studies}

\suppressfloats[t]
\paragraph{Contributions of the search design.}
Table~\ref{tab:ablation} cumulatively removes three components on AppWorld
Normal. We replace topic-guided selection with relevance ranking over the same
pool and budget, replace module-wise mutation with joint harness editing, and
finally remove research guidance to obtain Meta Harness. TGC and SGC decline
at every step for both backbones. Topic guidance alone contributes 1.4 TGC and
2.3 SGC points for Qwen, and 2.6 and 5.4 points for Mini, showing that topic
coverage adds value beyond relevance ranking.

\suppressfloats[t]
\begin{wraptable}{r}{0.5\textwidth}
  \vspace{-8pt}
  \centering
  \caption{Module composition on AppWorld DEV (57 tasks, three runs).
  Scores are mean $\pm$ SD (\%). N/A: module absent from the combination.}
  \label{tab:module_composition}
  \vspace{4pt}
  \fontsize{8}{9.5}\selectfont
  \setlength{\tabcolsep}{3pt}
  \renewcommand{\arraystretch}{1.12}
  \newcommand{\mcstd}[1]{{\fontsize{6}{7}\selectfont\color{gray}$\pm$#1}}
  \resizebox{\linewidth}{!}{%
  \begin{tabular}{lrrrr}
    \toprule
    & \multicolumn{2}{c}{\textbf{Qwen3.5-27B}}
    & \multicolumn{2}{c}{\textbf{GPT-5.4-mini}} \\
    \cmidrule(lr){2-3}\cmidrule(lr){4-5}
    Configuration & TGC $\uparrow$ & SGC $\uparrow$
      & TGC $\uparrow$ & SGC $\uparrow$ \\
    \midrule
    Initial harness & 67.8\mcstd{1.01} & 43.9\mcstd{3.04}
      & 66.1\mcstd{4.43} & 47.3\mcstd{9.12} \\
    $+$ Tool only & 72.5\mcstd{5.36} & 50.9\mcstd{13.25}
      & \multicolumn{2}{c}{\textcolor{gray}{N/A}} \\
    $+$ Context only & 74.9\mcstd{2.68} & 57.9\mcstd{5.26}
      & \multicolumn{2}{c}{\textcolor{gray}{N/A}} \\
    $+$ Skills only & 73.7\mcstd{7.02} & 47.4\mcstd{13.93}
      & 69.0\mcstd{3.65} & 52.6\mcstd{5.25} \\
    $+$ Memory only & 81.3\mcstd{1.01} & 59.6\mcstd{3.04}
      & 72.5\mcstd{2.02} & 45.6\mcstd{3.06} \\
    $+$ Workflow only & 77.2\mcstd{6.33} & 61.4\mcstd{10.96}
      & 72.5\mcstd{4.04} & 47.4\mcstd{10.55} \\
    \midrule
    \rowcolor{gray!10}
    Combined & \textbf{84.8}\mcstd{1.01} & \textbf{73.7}\mcstd{5.26}
      & \textbf{77.2}\mcstd{1.75} & \textbf{59.6}\mcstd{2.48} \\
    \bottomrule
  \end{tabular}%
  }
  \vspace{-12pt}
\end{wraptable}

\paragraph{Module interactions under crossover.}
Table~\ref{tab:module_composition} compares constituents and combinations
on AppWorld development tasks. Qwen combines all five modules, while Mini
combines skills, memory, and workflow, retaining the initial implementation
elsewhere. The combinations outperform their strongest constituent by
3.5 TGC/12.3 SGC points for Qwen and 4.7/7.0 for Mini. Module interactions also
change candidate rankings: replacing Qwen's 75.4\% standalone skills candidate
with a 73.7\% candidate raises the combination from 80.7\% to 84.8\%.
Standalone scores therefore do not identify the strongest composition.

\subsection{Analysis of Harness Evolution}
\label{sec:evolution_analysis}

\suppressfloats[t]
\begin{wraptable}{r}{0.48\textwidth}
  \vspace{-8pt}
  \centering
  \caption{Challenge TGC gains (points) by reference API count.}
  \label{tab:task_structure}
  \vspace{4pt}
  \scriptsize
  \setlength{\tabcolsep}{3.5pt}
  \renewcommand{\arraystretch}{0.92}
  \begin{tabular}{lccc}
    \toprule
    & \multicolumn{3}{c}{\textbf{Reference APIs}} \\
    \cmidrule(lr){2-4}
    \textbf{Backbone} & $\leq 9$ ($n$=168) & 10--12 ($n$=147) & $>12$ ($n$=102) \\
    \midrule
    Qwen3.5-27B & 13.1 & 18.4 & 9.5 \\
    GPT-5.4-mini & 5.4 & 15.9 & 7.2 \\
    \bottomrule
  \end{tabular}
  \vspace{-12pt}
\end{wraptable}
\paragraph{Generalization to broader API requirements.}
Table~\ref{tab:task_structure} groups Challenge tasks using boundaries from
training, where 76.7\% of tasks use at most nine APIs and the maximum is 12.
Both harnesses retain positive mean gains beyond this maximum, improving Qwen
and Mini by 9.5 and 7.2 TGC points. This result extends transfer to unseen
applications to broader API requirements. Appendix~\ref{app:api_distribution}
provides the complete distributions and confidence intervals.

\suppressfloats[t]
\begin{wrapfigure}{r}{0.50\textwidth}
  \vspace{-8pt}
  \centering
  \includegraphics[width=\linewidth]{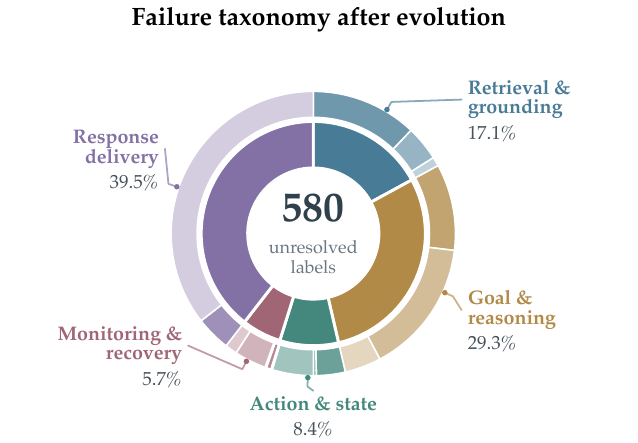}
  \vspace{-15pt}
  \caption{Residual Qwen3.5-27B failure taxonomy with five families and
  16 subtypes.}
  \label{fig:qwen_failure_taxonomy_overview}
  \vspace{-8pt}
\end{wrapfigure}
\paragraph{Residual failures after evolution.}
Auditing all 1,336 failed Qwen episodes yields five failure families and
16 subtypes. Figure~\ref{fig:qwen_failure_taxonomy_overview} summarizes 580
labels across 549 evolved-harness failures. Invalid response formats decline
from 492 to 206 episodes, while constraint violations (91 to 89) and incomplete
retrieval (77 to 70) change little. Evolution therefore improves answer
delivery most strongly, while requirement tracking and evidence coverage remain
recurring challenges. Appendix~\ref{app:failure_taxonomy} provides definitions,
annotation procedures, and complete frequencies.

\paragraph{Evolution expands task coverage and reliability.}
Figure~\ref{fig:evolution_summary} (right) pools Normal and Challenge while using
the same three runs as Table~\ref{tab:main}. Qwen improves on 221 of 585 tasks,
versus 57 regressions, while Mini improves on 200 tasks versus 106 regressions.
Among these gains, 73.3\% for Qwen and 66.0\% for Mini come from tasks that
succeeded intermittently under the initial harness. Qwen's tasks solved in all
three runs rise from 205 to 333. Evolution therefore improves reliability and
task coverage. Appendix~\ref{app:analysis_protocol} provides split-specific
matrices and further analysis.

\section{Conclusion}
We presented \textsc{ScholarEvolve}, a framework that evolves agent harnesses
around a fixed backbone by learning from research. It organizes exploration by
harness module, uses topic modeling to cover distinct mechanisms, and evaluates
module combinations before selection. The evolved harnesses improve both task
backbones on AppWorld and $\tau^2$-Bench and transfer to held-out AppWorld
applications, while new publications drive further gains in lifelong evolution.
Residual failures in requirement tracking and evidence coverage remain open
problems.

\subsubsection*{Acknowledgments}
The UCSB team acknowledges support from the National Science Foundation (NSF) under Grant Nos. 2338252, 2302730, and 2619240.

\bibliography{iclr2027_conference}
\bibliographystyle{iclr2027_conference}

\clearpage
\appendix
\etocdepthtag.toc{appendix}
\etocsettagdepth{mainmatter}{none}
\etocsettagdepth{appendix}{subsection}
\renewcommand{\contentsname}{Appendix Contents}
\tableofcontents
\clearpage

\section{Additional Experimental Details}

\subsection{Implementation Details}
\label{app:implementation_details}

\paragraph{Task environments and data.}
AppWorld uses the minimal official prompt, including the benchmark's
instruction block, and permits 100 environment steps per episode.
The 90 training tasks in $\mathcal D_{\mathrm{evo}}$ supply experience for
research auditing and offline artifact construction. The 57 development tasks
in $\mathcal D_{\mathrm{val}}$ determine candidate selection, whereas
$\mathcal D_{\mathrm{test}}^{\mathrm{N}}$ and
$\mathcal D_{\mathrm{test}}^{\mathrm{C}}$ evaluate the frozen harness.
Scenario variants remain grouped across these splits.
For Telecom, the 250-task experience pool is sampled with seed 2026 from
the full task collection after excluding the base collection. Stratified
sampling includes 18 service-issue, 60 mobile-data, and 172 MMS tasks.
The official 74-task training split forms $\mathcal D_{\mathrm{val}}$, and
$\mathcal D_{\mathrm{test}}$ is retained for reporting. The environment uses its
main policy and technical support manual, with a limit of 200 conversation
steps, ten errors, and simulation seed 300. Its user simulator is
GPT-5.2 with high reasoning effort.

\paragraph{Model configuration.}
The evolved task backbones are Qwen3.5-27B and GPT-5.4-mini, with a
maximum completion length of 65,536 tokens per model call. Mini uses low
reasoning effort. In the main experiments, Qwen is served by vLLM 0.27.1
with tensor parallelism four, temperature zero, and thinking enabled.
The Telecom server has a 262,144-token context limit. The archived
AppWorld requests include low reasoning effort, which the Qwen serving
template ignores while retaining its default thinking mode. Thus, the
server's thinking configuration determines Qwen's reasoning behavior.
Within each benchmark and backbone, initial and evolved harnesses use
the same task-model configuration and environment limits.

\paragraph{Search and artifact construction.}
Main-experiment research selection uses GPT-5.4 with medium reasoning
effort, and candidate code is
produced through Codex with GPT-5.4. The main AppWorld searches request
four paper-derived candidates per module, or 20 in total. Telecom requests
three per module, or 15. Single-module candidates are measured before
combinations are ranked and evaluated. AppWorld's initial development
measurements use three runs, with the Mini finalists additionally checked
over nine runs. Telecom development measurements use four internal trials
per task. The compared evolution approaches receive the same allocated
search rollout budget within each setting. Successful trajectories from
$\mathcal D_{\mathrm{evo}}$ can build persistent skill and memory artifacts,
which are frozen during evaluation on $\mathcal D_{\mathrm{val}}$ and
$\mathcal D_{\mathrm{test}}$. Runtime state can change within an episode.
The same AppWorld implementations and artifacts are used on Normal and
Challenge. For convergence based stopping, we set $s_{\mathrm{stop}}=1$:
one generation without a confirmed validation gain ends the current search
cycle. Lifelong settings are specified separately in
Appendix~\ref{app:lifelong}.

\paragraph{Metric aggregation.}
For each AppWorld run, TGC is the percentage of successful tasks and SGC
is the percentage of scenarios whose three variants all succeed. Normal
contains 56 scenarios and Challenge 139. We aggregate each metric over
three evaluation runs. For Telecom, every run contains four internal
trials per task. If task $i$ succeeds in $s_i$ of these trials, its
contribution to pass$^k$ is $\binom{s_i}{k}/\binom{4}{k}$, taken as zero
when $s_i<k$. We average over the 40 test tasks within a run, then report
the mean and standard deviation across three runs. The four-trial
estimator is computed separately in each run.

\paragraph{Ablation controls.}
The cumulative AppWorld ablation first substitutes relevance ranking for
cross-topic paper selection while preserving the screened pool and paper
budget. It then replaces isolated module mutation and recombination with
joint harness editing, and finally removes external research guidance.
For Qwen, the archived screened pool contains 1,956 module--paper records.
Upstream topic assignment and mechanism screening are held fixed when
changing selection. This comparison measures topic-guided selection
within that pool. The module-composition analysis separately compares
constituents with their exact combination on the same 57 development
tasks over three runs. Qwen combines all five modules, whereas Mini combines
skills, memory, and workflow while retaining the base tool and context
modules. These results describe development-set compositions. Qwen's combination
in Table~\ref{tab:module_composition} differs from its main test harness
in the skills implementation. The other four modules are identical.

\subsection{Lifelong Evolution Protocol and Results}
\label{app:lifelong}

Table~\ref{tab:lifelong} reports the best harness found in each generation,
selected on $\mathcal D_{\mathrm{val}}$ and evaluated on all 168 tasks in
$\mathcal D_{\mathrm{test}}^{\mathrm{N}}$. At each update,
\textsc{ScholarEvolve} selects at most one paper
per module from that round's publication window, builds up to five individual
mutations, and evaluates up to four module combinations on the 57 development
tasks. The best harness receives three development evaluations in total.
All three generations' search decisions are frozen before test evaluation.
The three disjoint publication windows are through December 2025,
January to April 2026, and May to August 2026. The windows use arXiv
first-submission dates and are reconstructed from the current index. Model
weights, archived $\mathcal D_{\mathrm{evo}}$ trajectories, and derived
artifacts remain fixed throughout. Research selection uses GPT-5.4 and candidate implementation
uses Codex GPT-5.5, both with medium reasoning effort.

\begin{table}[t]
  \centering
  \small
  \caption{Lifelong evolution on AppWorld Normal with Qwen3.5-27B.
  Scores are percentages. Here, $t_0$ is the shared initial harness, and
  $t_1$--$t_3$ report the best harness found in each generation.}
  \label{tab:lifelong}
  \setlength{\tabcolsep}{6pt}
  \begin{tabular}{@{}llcccc@{}}
    \toprule
    Metric $\uparrow$ & Method & $t_0$ & $t_1$ & $t_2$ & $t_3$ \\
    \midrule
    TGC & \textsc{ScholarEvolve} & $69.00\pm2.35$ & $73.41\pm2.09$ & $79.76\pm2.15$ & $\mathbf{81.55}\pm1.79$ \\
        & Meta Harness & $69.00\pm2.35$ & $71.03\pm2.81$ & $70.44\pm2.09$ & $69.64\pm2.73$ \\
    \midrule
    SGC & \textsc{ScholarEvolve} & $48.80\pm2.75$ & $54.17\pm2.72$ & $61.91\pm2.73$ & $\mathbf{66.67}\pm3.72$ \\
        & Meta Harness & $48.80\pm2.75$ & $47.62\pm2.73$ & $51.19\pm2.73$ & $51.19\pm2.06$ \\
    \bottomrule
  \end{tabular}
\end{table}

\subsection{Training and Test API Distributions}
\label{app:api_distribution}

\begin{figure}[t]
  \centering
  \includegraphics[width=\linewidth]{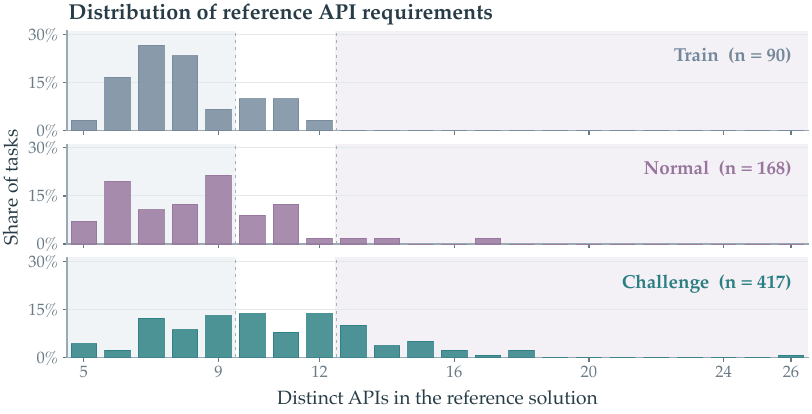}
  \caption{\textbf{Reference API distributions across AppWorld splits.}
  Bars show the percentage of tasks at each integer API count, with shared
  axes across the complete splits. The shaded regions mark at most nine
  APIs and more than 12 APIs, using the training third quartile and maximum.
  Dashed lines separate these ranges at integer-bin boundaries.}
  \label{fig:api_distribution}
\end{figure}

Figure~\ref{fig:api_distribution} reports the complete distributions of
reference API counts for Train, Normal, and Challenge. Each count records
the number of distinct APIs required by the official reference solution,
including helper APIs. It describes API breadth in that solution. Agent
interaction lengths are measured separately. The training distribution has
a third quartile of nine APIs and a maximum of 12. These thresholds define
the three groups in Table~\ref{tab:task_structure}, independently of test
gains. They contain 168, 147, and 102 Challenge tasks, respectively.
The API-count median increases from eight in Train to 8.5 in Normal and
ten in Challenge. Of the Challenge tasks, 24.5\% exceed the training maximum.
The reference API-count maximum in $\mathcal D_{\mathrm{val}}$ is ten.

The gains in Table~\ref{tab:task_structure} use the same three saved runs
per harness as Table~\ref{tab:main}. Their 95\% intervals use 20,000 paired
scenario bootstrap resamples with seed 20260922, preserving the three task
variants within each scenario and conditioning on the saved runs.
For the group above the training maximum, Qwen's interval is
$[-0.7,19.6]$ percentage points and Mini's is $[0.3,14.1]$.

\subsection{Analysis Protocol}
\label{app:analysis_protocol}

\begin{figure}[t]
  \centering
  \includegraphics[width=\linewidth]{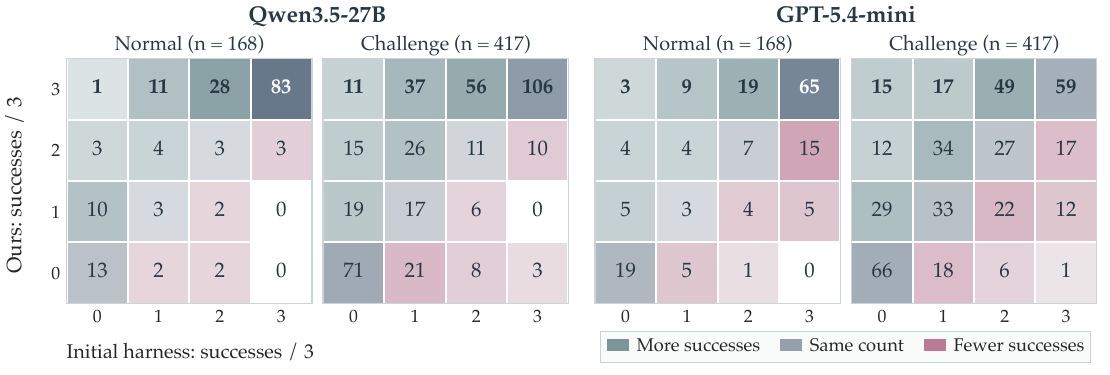}
  \vspace{-15pt}
  \caption{\textbf{Reliability transitions by AppWorld test split.}
  Columns and rows give initial and evolved success counts out of three.
  Cells report task counts. Teal denotes more successes, slate the same
  count, and rose fewer successes. Darker shades indicate larger fractions
  of each split on a shared square-root scale.}
  \label{fig:reliable_coverage_full}
\end{figure}

Task-level analyses use the final configurations reported in
Table~\ref{tab:main}. AppWorld subgroup gains are recomputed from native
success bits. Its difficulty labels come from benchmark metadata.
The training-referenced intervals follow Appendix~\ref{app:api_distribution}.
Other subgroup intervals use 20,000 paired cluster bootstrap resamples with seed
20260905: whole scenarios are resampled for AppWorld, preserving their
three related task variants, and whole tasks for Telecom. Intervals condition
on the recorded runs and describe exploratory subgroup comparisons.
Successful-run comparisons pair tasks across harnesses. Run indices
are not assumed to represent matched random seeds. The split-specific heatmaps
in Figure~\ref{fig:reliable_coverage_full} include
every task in both official test splits: 168 Normal and 417 Challenge
tasks. The archived task IDs and five difficulty/structure metadata fields
match the official AppWorld release for all 585 tasks. For Qwen, the
numbers of tasks with higher, equal, and lower success counts are
57/102/9 on Normal and 164/205/48 on Challenge. For Mini, they are
44/94/30 and 156/185/76. Proportions of newly solved and more frequently
solved tasks condition on a higher success count. Color intensity uses
the fraction of all tasks, with a common square-root mapping over
0--50\%. Mini Challenge uses the same three initial-harness reroll
evaluations and three final-harness runs as the main table.
Differences of a few
hundredths of a percentage point from the main table arise from computing
exact task means instead of aggregating rounded native scores.

\paragraph{Reliability transitions.}
Among tasks with an increased success count, the fractions that initially
succeeded in one or two runs are 75.4\% and 72.6\% for Qwen on Normal
and Challenge, and 72.7\% and 64.1\% for Mini. Tasks solved in all three
runs increase from 86 to 123 and from 119 to 210 for Qwen, and from 85
to 96 and from 89 to 140 for Mini. Of Qwen's newly consistent tasks,
39 of 40 on Normal and 93 of 104 on Challenge previously succeeded
in one or two runs.

\paragraph{Recovery after execution errors.}
Successful tool use includes recovery from intermediate mistakes.
In Qwen's Normal evaluation, 249 of 504 evolved-harness episodes encounter
an execution error, yet 180 of them ultimately succeed. The success rate
among these episodes is 72.3\%, compared with 64.5\% for the initial
harness. On Challenge, the corresponding rates are 59.9\% and 44.9\%.
These trajectory statistics show why an intermediate tool error provides
an incomplete account of an agent's ability: many successful tasks require
using the returned error to revise subsequent actions. The task outcome
and the path taken after an error together characterize execution robustness.

An execution error is identified by the benchmark's explicit
\texttt{Execution failed. Traceback:} sentinel. Recovery statistics report
final success among episodes that encounter this sentinel. The initial
and evolved harnesses can encounter errors on different tasks, so these
conditional rates describe their realized trajectories. On Normal the
success counts are 176 of 273 error-exposed episodes for the initial
harness and 180 of 249 for the evolved harness. On Challenge, they are
405 of 903 and 518 of 865, respectively. The qualitative mechanism map
is constructed from candidate implementations and depicts alternative
mechanisms within modules.

\subsection{Failure Taxonomy and Diagnostic Checks}
\label{app:failure_taxonomy}

\begin{figure}[t]
  \centering
  \includegraphics[width=\linewidth]{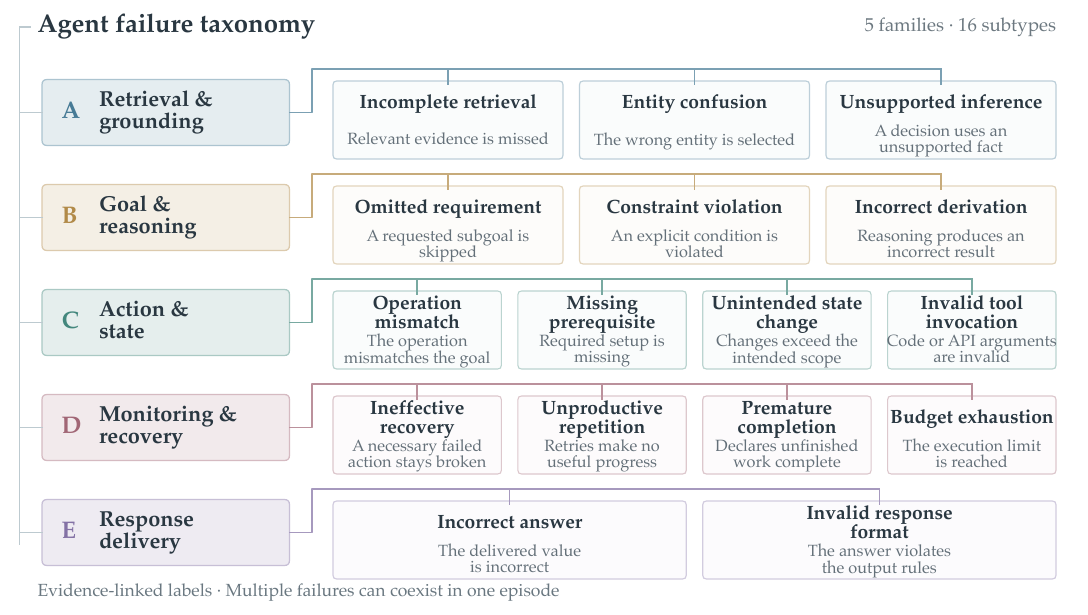}
  \caption{\textbf{A hierarchy of observable agent failures.}
  Five families organize evidence-linked deviations found in Qwen3.5-27B
  AppWorld trajectories. Each subtype identifies a distinct behavior or
  execution outcome. Labels can coexist within an episode. Recovery status
  is recorded separately.}
  \label{fig:qwen_failure_taxonomy}
\end{figure}

Figure~\ref{fig:qwen_failure_taxonomy} expands the taxonomy summarized in
Figure~\ref{fig:qwen_failure_taxonomy_overview} with a definition for each subtype.

The failure audit uses the complete Qwen3.5-27B initial/final cohorts:
168 Normal tasks and 417 Challenge tasks, with three executions per task
and harness. We verify the archived API requests and execution statuses
against their recorded hashes and match every task's success count to
the main-result cohort. The original evaluator outcomes remain fixed.
Released reference answers and evaluation definitions are used only for
this post-hoc diagnosis.

\paragraph{Hierarchical behavior coding.}
We annotate the 1,336 unsuccessful episodes, comprising 787 initial-harness
and 549 evolved-harness failures across 396 tasks. A stratified calibration
set contains 50 episodes from distinct scenarios, spanning both splits,
both harnesses, and the endpoint checks below. Fourteen targeted rechecks
refine ambiguous boundaries before the codebook is frozen. The full cohort,
including calibration episodes, is then processed with the frozen definitions.
The annotator is Muse Spark 1.3 Contributor via OpenRouter, with temperature
zero and medium reasoning effort. Inputs contain the task, shared base
instructions, complete executed code and returned observations, and released
reference answers and evaluation definitions. Model and harness identities,
comparison groups, trial numbers, earlier diagnostic labels, and performance
gains are hidden. Identity and credential fields receive consistent
within-episode pseudonyms. Duplicate raw model text is excluded, and no executed
steps are truncated.

A second pass receives the full input, draft labels, and reconstructed
completion and answer checks. It revisits each proposed deviation and
its subsequent recovery, retaining literal evidence quotes and step indices.
Both passes use the same model, and the second sees the first-pass draft.
Local checks match evidence quotes to source text and
reconcile labels with verified completion signals, execution limits, and
answer rules. All corrections preserve the original model output and
record their rationale. A targeted review restores locally verified
relationships between full addresses and street components that receive
different pseudonyms. These alias differences do not establish an incorrect
location. No original address values are sent to the annotator.

Each issue is marked unresolved, recovered, or uncertain, with an evidence
strength and confidence rating. Supported unresolved counts require
medium or high confidence and directly observed evidence or a verified
reference comparison. An inference that depends on an unobserved condition
remains uncertain. Categories can overlap: a derivation error may produce
an incorrect answer, and an invalid completion response may accompany an
unmet state requirement. Conversely, a corrected tool invocation remains
recovered even when another issue causes the episode to fail. We do not
assign omitted requirements solely because a progressing run reaches its
step budget. Missing per-check outcomes or final state snapshots remain
unobserved. These labels characterize archived behavior under the evaluated
harnesses and search budget.

\paragraph{Coverage and descriptive comparisons.}
Sixteen of the 17 candidate subtypes have supported instances, while the
missing-answer subtype is unobserved and omitted from both taxonomy figures.
There are 218 failed episodes without a supported unresolved behavior
label, including 97 initial-harness and 121 evolved-harness episodes.
Their uncertainty is retained. Table~\ref{tab:hierarchical_failures}
reports subtype counts. The response-format subtype includes one initial
Normal episode with an otherwise correct quotation and extraneous
attribution, in addition to the null-answer violations below.

We also group tasks by the supported unresolved labels in their initial
executions and compare their successful-run counts across harnesses.
Among tasks initially failing all three runs and exhibiting an invalid
response format, 54 of 111 achieve at least one success after evolution.
For tasks exhibiting constraint violations, the corresponding count is
16 of 42. Groups can overlap and include other co-occurring deviations.
These are task-level transitions conditional on observed initial behavior.
The archived labels and task counts provide the complete cross-tabulation.

\begin{table}[t]
\centering
\small
\caption{Supported unresolved deviations in the full Qwen AppWorld cohorts. Counts are multi-label. Each harness has 504 Normal and 1,251 Challenge episodes. The final column counts recovered occurrences within the failed-episode cohort, pooled across splits and harnesses.}
\label{tab:hierarchical_failures}
\begin{tabular}{lrrrrr}
\toprule
& \multicolumn{2}{c}{Normal} & \multicolumn{2}{c}{Challenge} & Recovered \\
Subtype & Initial & Evolved & Initial & Evolved & (pooled) \\
\midrule
\multicolumn{6}{l}{\textit{Retrieval \& grounding}} \\
Incomplete retrieval & 12 & 8 & 65 & 62 & 0 \\
Entity confusion & 7 & 11 & 8 & 12 & 0 \\
Unsupported inference & 1 & 1 & 3 & 5 & 0 \\
\multicolumn{6}{l}{\textit{Goal \& reasoning}} \\
Omitted requirement & 6 & 12 & 44 & 45 & 0 \\
Constraint violation & 30 & 31 & 61 & 58 & 5 \\
Incorrect derivation & 13 & 11 & 11 & 13 & 1 \\
\multicolumn{6}{l}{\textit{Action \& state}} \\
Operation mismatch & 7 & 5 & 8 & 14 & 1 \\
Missing prerequisite & 2 & 0 & 1 & 2 & 49 \\
Unintended state change & 2 & 1 & 34 & 26 & 3 \\
Invalid tool invocation & 0 & 0 & 0 & 1 & 89 \\
\multicolumn{6}{l}{\textit{Monitoring \& recovery}} \\
Ineffective recovery & 1 & 0 & 4 & 3 & 0 \\
Unproductive repetition & 2 & 0 & 3 & 1 & 0 \\
Premature completion & 6 & 3 & 24 & 18 & 0 \\
Budget exhaustion & 1 & 0 & 8 & 8 & 0 \\
\multicolumn{6}{l}{\textit{Response delivery}} \\
Incorrect answer & 14 & 6 & 11 & 17 & 0 \\
Invalid response format & 85 & 29 & 407 & 177 & 0 \\
\bottomrule
\end{tabular}
\end{table}

\paragraph{Completion and answer diagnostics.}
An additional, mutually exclusive diagnostic assigns each failed episode
the first failed endpoint check in an ordered procedure.
First, absence of the recorded completion signal yields
\textit{execution completion}. Second, a completed action task whose
reference answer is null receives \textit{response contract} if its
submitted answer does not normalize to null. Third, a completed task
requiring an answer receives \textit{answer correctness} if that answer
fails its published comparison rule. Finally, an unsuccessful episode
with a completed interaction and an accepted answer receives
\textit{goal-state realization}: at least one remaining environment
requirement is unsatisfied. These endpoint labels locate observable
failures. Information retrieval, planning, state tracking, and local
reasoning can contribute to the same endpoint.

We apply the published answer normalization and the task-specific rules
for numerical tolerances, unordered lists, acceptable quotations, and
sets of returned codes. All successful episodes pass the reconstructed
completion and answer checks. Category incidence uses all episodes as
the denominator, 504 or 1,251 per harness. Total failures decrease from
156 to 94 on Normal and from 631 to 455 on Challenge. In diagnostic
order, category counts change from $(1,84,15,56)$ to $(0,29,8,57)$ on
Normal and from $(10,407,11,203)$ to $(10,177,17,251)$ on Challenge.
The ordered labels can expose a later failure once an earlier check
passes, so the increase in a later category combines changes in
execution with changes in which failure is recorded first.

\begin{figure}[t]
  \centering
  \includegraphics[width=\linewidth]{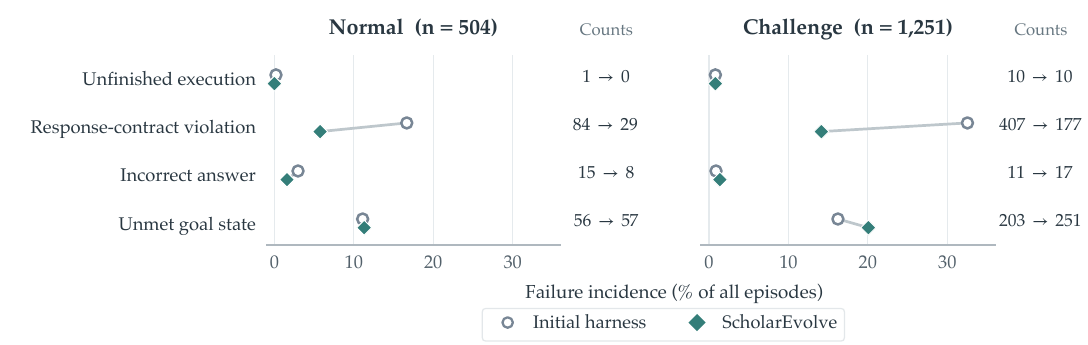}
  \caption{\textbf{Completion and answer diagnostics on the full Qwen cohorts.}
  Points show fractions of all episodes, and counts give initial $\rightarrow$
  evolved totals. The four endpoint categories follow the diagnostic order
  from top to bottom. Behavior labels in Figure~\ref{fig:qwen_failure_taxonomy}
  provide a separate, overlapping description of the trajectories.}
  \label{fig:qwen_endpoint_diagnostics}
\end{figure}

The qualitative state examples are tasks \texttt{9126bf0\_3} and
\texttt{20c1328\_3}. All three initial and all three evolved traces are
checked for each example. The first distinguishes creation from update,
whereas the second checks whether pre-existing cart contents enter an order.
These cases explain concrete mechanisms, while the frequency estimates
use the complete test cohorts. Residual failures describe the final
harness under the evaluated search budget and provide targets for
subsequent harness refinement or targeted training data.

\section{Qualitative Analysis}
\label{sec:qualitative_analysis}

\begin{figure}[t]
  \centering
  \includegraphics[width=\linewidth]{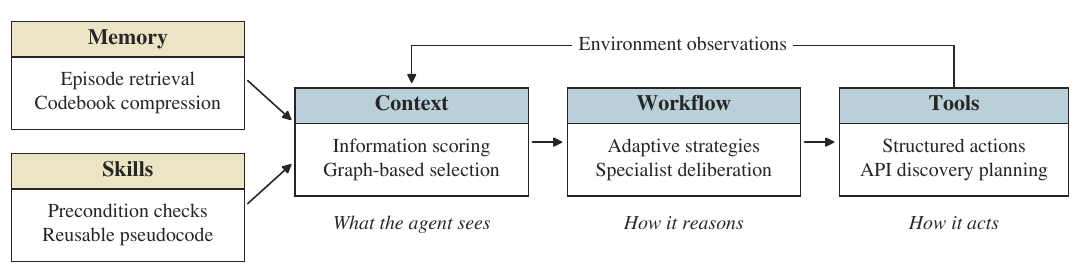}
  \caption{\textbf{Implemented mechanisms span the agent's decision process.}
  Schematic organization of alternatives from the Qwen AppWorld candidate
  pool. Each module lists two research-derived mechanisms. Arrows show
  information and action flow through the harness. The alternatives are
  candidates for selection and recombination.}
  \label{fig:mechanism_map}
\end{figure}

\paragraph{Research introduces choices at different decision points.}
Figure~\ref{fig:mechanism_map} maps implemented Qwen AppWorld candidates
onto the agent's execution cycle. The alternatives operate on what the
agent sees, how it reasons, and how it expresses an action.
For example, one context candidate prioritizes text by information content,
while another scores relationships among context segments before selecting
them under a length budget. Both act on the evidence available for the
next decision, but embody different selection mechanisms. Memory candidates
retrieve summaries of solved episodes or compress experience into a
bounded codebook. Skill candidates provide procedures that check their
preconditions or compile recurring API sequences into reusable pseudocode.
These changes expose reusable knowledge at execution time, complementing
workflow candidates that route among reasoning strategies or aggregate
specialist proposals. The resulting search explores multiple sources of
agent improvement, including information selection, experience reuse,
action representation, and deliberation.

\paragraph{State-aware execution turns some repeated failures into successes.}
The Qwen trajectories illustrate two reusable mechanisms within the
action and state family. In one task, the initial harness creates a new object
when the request requires updating an existing one, whereas the evolved harness
locates and modifies the original object. In another, an operation
unintentionally includes pre-existing state, whereas the evolved harness clears
that state before executing the requested operation. Both tasks move
from zero to three successful runs. These examples highlight object
identity and action preconditions as useful targets for harness design.
Residual failures also retain concrete intervention points: among the
71 Challenge tasks that fail in all three runs under both harnesses,
35 still exhibit an invalid final response in at least one evolved
run. They motivate further work on state checks and interface enforcement.

\section{Additional Method Details}
\label{app:method_details}

\subsection{Representative Module Implementations}
The following implementations illustrate the module interfaces introduced
in Section~\ref{sec:modules}.

\paragraph{Tool interface.}
This module exposes available actions and translates a model proposal into an
environment action. An AppWorld mutation introduces documentation lookup macros
that expand into executable API inspection calls. It also extracts Python
actions, checks their syntax and statement structure, applies bounded
normalization rules, and supports recovery through documentation queries.
This places action formatting and recovery at the boundary
between reasoning and execution.

\paragraph{Context management.}
This module assembles the model input from the policy, interaction history,
retrieved memories, and selected skills. An AppWorld mutation scores context
segments by task relevance and recency, reserves a minimum allocation for
selected segments, and distributes the remaining character budget according
to these scores. It then extracts query focused passages and restores their
chronological order while preserving the host policy. The resulting mutation
jointly controls information selection, compression, and presentation.

\paragraph{Skills.}
This module supplies reusable procedures constructed from
$\mathcal D_{\mathrm{evo}}$ trajectories and matched to the current task. An
AppWorld mutation mines
recurring API call subsequences from successful trajectories and compiles them
into pseudocode procedures with input and output fields, applicability cues,
ordered calls, and empirical support counts. At runtime, lexical trigger
matches and support counts select a procedure for context management, making
repeated patterns of tool use available as explicit plans.

\paragraph{Memories.}
This module retrieves historical experience relevant to the current task.
A Telecom mutation converts $\mathcal D_{\mathrm{evo}}$ episodes into
structured scene records with entity profiles, temporal attributes, task
attributes, and trajectory summaries. It retrieves candidate scenes through
entity matches, ranks them using weighted attribute and lexical overlap, and
returns structured evidence with a compact account of the earlier interaction.
This representation retains the concrete circumstances and outcomes of past
episodes.

\paragraph{Agentic workflows.}
This module organizes the model calls that produce the next action proposal.
An AppWorld mutation obtains separate proposals from planner, solver, and
critic roles, then invokes a verifier to select an action. Bounded, episode
local banks of recent observations and error conditioned retry guidelines
inform subsequent steps. The workflow returns a single proposal with aggregate
model usage, exposing its decision to the tool interface for execution.

\subsection{Research Pool Construction and Topic Selection}
\label{app:research_pipeline}

The pipeline connects observed capability gaps to research mechanisms
through the following stages, corresponding to the audit, retrieval,
and topic modeling stages in Figure~\ref{fig:ripe_method}.

\paragraph{Failure audit and capability abstraction.}
When an audit is available, we inspect $\mathcal D_{\mathrm{evo}}$ trajectories and group
failures by their observed causes. Audit records include fault attribution,
a cause description, and supporting observations. We prioritize recurring
agent-attributable failures judged amenable to changes at inference time.
The research model receives these summaries and representative evidence,
then produces a brief containing a general capability gap and a problem
summary. It removes benchmark names, domain entities, and particular tool
or field names from the research questions. For example, losing track of
an earlier observation can motivate research on context selection,
state tracking, or memory retrieval.

\paragraph{Module specific query construction.}
For each of the five modules, the research model receives the same brief,
the module's responsibility and interface, and the current harness behavior.
It describes how a change within that module could address the capability
gap, then proposes two query groups: broad phrases naming the capability
and narrower phrases naming potential mechanisms. We interleave the groups
to cover both problem areas and candidate solutions. Queries use short,
established research terms, such as \emph{context compression},
\emph{trajectory retrieval}, and \emph{skill discovery}. Each module
receives its own queries because a single capability gap can admit
interventions at several points in the harness. Module based query sets
also support research collection without an audit.

\paragraph{Retrieval and pool construction.}
We retrieve papers through arXiv using the generated query phrases and
collect their titles, abstracts, and source links. Within each module,
we deduplicate results by normalized title and exclude entries without
an abstract. A paper can remain in several module pools when its mechanisms
are relevant to different responsibilities. The resulting $\mathcal P_j$
is the input corpus for topic modeling. The implementation budget $K_j$
is applied after the topic taxonomy has been constructed.

\paragraph{Mechanism taxonomy and evidence assignment.}
The research model reads batches of titles and abstracts alongside the
taxonomy accumulated so far. Prompts request short mechanism names and
one-sentence descriptions, reuse existing names where appropriate, and
introduce new categories for uncovered mechanisms. Scope instructions
focus on implementable mechanisms for language model agents and exclude
surveys, standalone benchmarks, and unrelated uses of the same vocabulary.
Refinement merges synonymous or overlapping categories, absorbs overly
specific variants, and standardizes the granularity of the remaining
topics. The assignment stage then receives this fixed taxonomy and assigns
each paper an ordered list of supported topics, with an abstract excerpt
for each assignment. The parser retains recognized labels, and the first
retained label determines the paper's selection cluster.

\paragraph{Mechanism screening and candidate allocation.}
We compare each topic's mechanism with a description of the current
module and backbone capabilities, recording whether it would add an
absent mechanism and the reason for that decision. Topics already
subsumed by the current harness are removed from candidate allocation.
The remaining nonempty clusters are ordered by size, and papers are
selected in round robin order until $K_j$ candidates have been allocated
or the eligible pool is exhausted. Retrieval order is preserved within
each cluster. This separates corpus collection from the diversity control
used to spend the implementation budget. Topic descriptions, merge
records, assignment excerpts, screening decisions, and selected paper
identities preserve the provenance of the resulting mutation directions.

\subsection{Module Interfaces and Executable Checks}
\label{app:module_checks}

The interfaces specify how module outputs enter execution. Memory modules
return evidence, and skill modules return procedures. Context management
places these outputs in the model input. Workflow modules receive that
input and preserve it as a prefix of their model calls, appending transient
deliberation when needed. They return action proposals for the tool
interface to execute. Modules may read shared episode state, while these
output contracts preserve the responsibility of each module during
mutation and crossover.

Before measurement,
executable health probes check importability, interface compatibility,
persistent artifact construction and reloading, and valid action production
with controlled model responses. The coding agent repairs reported errors
until these checks pass or the repair budget is exhausted.

Mutations can construct persistent artifacts from successful trajectories in
$\mathcal D_{\mathrm{evo}}$, such as the procedure libraries and scene memories
described above. Each artifact is saved with its source-task provenance and
remains fixed during evaluation on $\mathcal D_{\mathrm{val}}$ and
$\mathcal D_{\mathrm{test}}$.
Modules can maintain local state within an evaluation episode as new actions
and observations become available.

\subsection{Crossover Diagnostics}
Each shortlisted configuration is assembled through the module interfaces and
evaluated as a complete harness $H_c$. Its observed gain and deviation from
the screening prediction are
\begin{equation}
\begin{aligned}
    \Delta(c)&=\frac{1}{|\mathcal D_{\mathrm{val}}|}
       \sum_{i\in\mathcal D_{\mathrm{val}}}
       \left[\bar r_i(H_c)-\bar r_i\!\left(H^{(0)}\right)\right],&
    \varepsilon(c)=\Delta(c)-\widehat\Delta(c).
\end{aligned}
\label{eq:composition_residual}
\end{equation}
The additive score can exceed the gain attainable under bounded rewards.
Joint evaluation supplies the performance used for selection. The residual
captures deviations arising from module interactions, reward ceilings, and
evaluation variation. We rank measured configurations by their validation
gain and can evaluate subsets of the strongest individual mutations to
identify useful inclusions or omissions. A selected configuration, together
with its frozen artifacts, becomes a reusable harness for further search.

The lifelong loop applies the retention rule in
Equation~\ref{eq:champion_update} to the best evaluated mutation or
crossover candidate. The selected harness and its artifacts are fixed before
held out test evaluation.

\clearpage
\section{Candidates and Evolved Harnesses}
\label{app:harness_code}

\subsection{Candidate Pool for Qwen3.5-27B on AppWorld}
\label{app:candidate_pool}

Table~\ref{tab:all_candidates} lists all 20 paper-derived candidates from the
main Qwen3.5-27B AppWorld search, four per module. Within each module, the four
source papers come from four distinct topics. Seventeen candidates raise
standalone development TGC, and eight have a 90\% interval above zero. Three
candidates reduce it sharply, and the final harness excludes them.

\begin{table}[!h]
\centering
\scriptsize
\caption{All candidates for Qwen3.5-27B on AppWorld. Each candidate changes one
module of the initial harness. $\Delta$ is the standalone TGC gain in points on
the 57 development tasks over three runs, with a 90\% paired task-bootstrap
interval. W/L counts tasks whose mean reward is higher/lower than under the
initial harness. $\star$ marks the modules in the final harness.}
\label{tab:all_candidates}
\vspace{2pt}
\setlength{\tabcolsep}{4pt}
\renewcommand{\arraystretch}{1.08}
\resizebox{\linewidth}{!}{%
\begin{tabular}{@{}lllrcrc@{}}
\toprule
Module & Topic & Source paper & $\Delta$ & 90\% interval & W/L & \\
\midrule
\multirow{4}{*}{Tool interface} & Action Space Control & Budget-constrained tool learning~\citep{zheng2024budget} & +2.9 & [$-$3.5, +9.4] & 15/11 &  \\
 & Abstract Strategy Actions & CheMatAgent~\citep{wu2025chematagent} & +1.2 & [$-$4.1, +6.4] & 12/10 &  \\
 & Structured API Actions & ToolACE-R~\citep{zeng2026toolace} & +4.7 & [$-$0.6, +9.9] & 14/8 & $\star$ \\
 & Constructed Composite Actions & RefTool~\citep{liu2026reftool} & +1.8 & [$-$4.7, +8.2] & 15/10 &  \\
\midrule
\multirow{4}{*}{Context} & Lossy Context Rewriting & Compression without MLPs~\citep{honig2025better} & $-$45.6 & [$-$53.8, $-$36.8] & 1/40 &  \\
 & Independent Unit Scoring & ICPC~\citep{yu2025icpc} & $-$43.3 & [$-$52.0, $-$34.5] & 2/40 &  \\
 & Budget and Scope Routing & Prompt compression limits~\citep{nagle2024fundamental} & +7.0 & [+1.8, +12.3] & 16/7 & $\star$ \\
 & Global Subset Selection & Prompt-SAW~\citep{ali2024prompt} & +4.1 & [$-$1.8, +9.9] & 16/11 &  \\
\midrule
\multirow{4}{*}{Skills} & Skill Induction & RL with skill library~\citep{wang2026reinforcement} & +2.9 & [$-$2.9, +8.8] & 12/10 &  \\
 & Skill Selection & Group of Skills~\citep{zeng2026group} & +0.6 & [$-$6.4, +7.6] & 15/15 &  \\
 & Skill Maintenance & Skill Drift Is Contract Violation~\citep{fan2026skill} & +5.8 & [+1.2, +10.5] & 15/5 &  \\
 & Skill Compilation & Skill-as-Pseudocode~\citep{li2026skill} & +7.6 & [+1.2, +14.0] & 15/7 & $\star$ \\
\midrule
\multirow{4}{*}{Memories} & Structured Memory Architecture & Oracle Agent Memory~\citep{alake2026oracle} & +13.5 & [+8.2, +19.3] & 20/3 & $\star$ \\
 & Memory Writing and Maintenance & Memory bank compression~\citep{katraouras2026memory} & +4.1 & [$-$2.9, +11.1] & 16/12 &  \\
 & Retrieval Scoring and Routing & Episodic-memory prompting~\citep{do2024large} & +9.9 & [+4.1, +15.8] & 16/5 &  \\
 & Adaptive Memory Optimization & CoEvo-Mem~\citep{ye2026coevo} & $-$67.8 & [$-$75.4, $-$60.2] & 0/50 &  \\
\midrule
\multirow{4}{*}{Workflow} & Evaluator-Guided Refinement & Perceptual self-reflection~\citep{shende2026perceptual} & +8.8 & [+2.3, +15.2] & 17/8 &  \\
 & Candidate Aggregation & CodeMonkeys~\citep{ehrlich2025codemonkeys} & +8.2 & [+2.3, +14.0] & 17/8 &  \\
 & Adaptive Compute Routing & RTTC~\citep{munoz2025rttc} & +4.1 & [$-$1.8, +10.5] & 14/12 &  \\
 & Multi-Agent Deliberation & TMAS~\citep{wu2026tmas} & +9.4 & [+3.5, +15.8] & 17/6 & $\star$ \\
\bottomrule
\end{tabular}}
\end{table}

\subsection{Selected Configurations and Code}
\label{app:final_code}

Table~\ref{tab:final_configs} lists the implementation selected for each module
in the four main-experiment harnesses, with the source paper and the topic
assigned during topic modeling. All selections were fixed on
$\mathcal D_{\mathrm{val}}$ before test evaluation. The listings below show
excerpts of the five modules in the final Qwen3.5-27B AppWorld harness
reported in Table~\ref{tab:main}, with omitted lines marked. Each module
implements the interface in Appendix~\ref{app:module_checks}. Modules with
persistent artifacts build them from $\mathcal D_{\mathrm{evo}}$ in
\texttt{fit} and reload them at evaluation time.

\begin{table}[!h]
\centering
\scriptsize
\caption{Module implementations in the main-experiment harnesses. Each entry
gives the source paper and, in italics, the assigned topic. A dash retains the initial module.}
\label{tab:final_configs}
\vspace{2pt}
\setlength{\tabcolsep}{3pt}
\renewcommand{\arraystretch}{1.25}
\begin{tabular}{@{}l>{\raggedright\arraybackslash}p{0.205\linewidth}>{\raggedright\arraybackslash}p{0.205\linewidth}>{\raggedright\arraybackslash}p{0.205\linewidth}>{\raggedright\arraybackslash}p{0.205\linewidth}@{}}
\toprule
& \multicolumn{2}{c}{\textbf{AppWorld}} & \multicolumn{2}{c}{\textbf{$\tau^2$-Bench Telecom}} \\
\cmidrule(lr){2-3}\cmidrule(lr){4-5}
Module & Qwen3.5-27B & GPT-5.4-mini & Qwen3.5-27B & GPT-5.4-mini \\
\midrule
Tool interface
 & ToolACE-R~\citep{zeng2026toolace} \textit{Structured API Actions}
 & --
 & Lower Privileges Suffice~\citep{yang2026lower} \textit{Action-Centric Post-Training}
 & Interactive Semantic Parsing~\citep{yao2019model} \textit{Action Validation and Repair} \\
Context
 & Prompt compression limits~\citep{nagle2024fundamental} \textit{Budget and Scope Routing}
 & --
 & Diable~\citep{lesci2023diable} \textit{Structured State Memory}
 & Measure Before You Manage~\citep{chen2026measure} \textit{Context Buffer Management} \\
Skills
 & Skill-as-Pseudocode~\citep{li2026skill} \textit{Skill Compilation}
 & Skill Drift Is Contract Violation~\citep{fan2026skill} \textit{Skill Library Maintenance}
 & Procedural KG Extraction~\citep{carriero2024human} \textit{Skill Induction}
 & Procedural Knowledge at Scale~\citep{wu2026procedural} \textit{Procedural Memory and Retrieval} \\
Memories
 & Oracle Agent Memory~\citep{alake2026oracle} \textit{Structured Memory Architecture}
 & CoEvo-Mem~\citep{ye2026coevo} \textit{Adaptive Retrieval Control}
 & CAST~\citep{ma2026cast} \textit{Structured Relational Memory}
 & MemRL~\citep{zhang2026memrl} \textit{Adaptive Retrieval and Retention} \\
Workflow
 & TMAS~\citep{wu2026tmas} \textit{Multi-Agent Deliberation}
 & Plan-and-Solve~\citep{wang2023plan} \textit{Plan-and-Execute}
 & Textual-to-Visual Self-Verification~\citep{xu2025textual} \textit{Hierarchical Decomposition}
 & Same-Model Self-Verification~\citep{phalod2026should} \textit{Adaptive Routing} \\
\bottomrule
\end{tabular}
\end{table}
\FloatBarrier

\begin{harnesscode}{Tool interface: ToolACE-R {\mdseries\itshape (Structured API Actions)}}{tool-p3.py}
# ... lines 1-16 omitted ...
DOC_ACTION = ToolSpec(
    name="DOC_ACTION",
    description=(
        "DOC_ACTION <app> or DOC_ACTION <app>.<api> expands into Python that prints the "
        "available app APIs or one API document before the next reasoning step."
    ),
    parameters={"target": "str"},
    composite_of=("python_code",),
)
    # ... lines 26-77 omitted ...
    def act(self, proposal, tools=None, state=None, task=None, client=None):
        del tools, task, client
        candidates = self._proposal_candidates(proposal)
        budgeted_rounds = self.max_refinement_rounds
        if state is not None:
            budgeted_rounds = min(
                budgeted_rounds,
                max(1, min(3, int(getattr(state, "steps_left", budgeted_rounds) or budgeted_rounds))),
            )
        for candidate in candidates:
            refined = candidate
            for _ in range(budgeted_rounds):
                if self._is_effectful_python(refined):
                    return refined
                updated = self._refine_once(refined, proposal=proposal, state=state)
                if updated == refined:
                    break
                refined = updated
            if self._is_effectful_python(refined):
                return refined
        return self._fallback_action(proposal=proposal, state=state)
    # ... lines 99-196 omitted ...
    def _doc_action(self, app, api):
        if app and api:
            return f"print(apis.api_docs.show_api_doc(app_name={app!r}, api_name={api!r}))"
        if app:
            return f"print(apis.api_docs.show_api_descriptions(app_name={app!r}))"
        return "print(apis.api_docs.show_app_descriptions())"
    # ... lines 203-212 omitted ...
    def _is_effectful_python(self, code):
        normalized = self._normalize_code(code)
        if not normalized:
            return False
        try:
            tree = ast.parse(normalized)
        except SyntaxError:
            return False
        if not tree.body:
            return False
        for node in tree.body:
            if isinstance(node, ast.Expr):
                if isinstance(node.value, ast.Call):
                    return True
                continue
            return True
        return False
# ... lines 230-236 omitted ...
\end{harnesscode}

\begin{harnesscode}{Context management: Prompt compression limits {\mdseries\itshape (Budget and Scope Routing)}}{cont-p3.py}
"""Query-aware, variable-rate context compression for the context_management slot.

The paper's key result for this slot is that compression must depend on the downstream query.
This implementation follows that spirit with an Adaptive QuerySelect-style policy:

* preserve the host policy;
* score candidate context blocks by task relevance plus recency;
* keep blocks in their original order;
* allocate more characters to more relevant blocks under the advertised budget.
"""
        # ... lines 11-159 omitted ...
        for index, turn in enumerate(history):
            content = _content(turn)
            if not content:
                continue
            relevance = self._score_text(content, query_terms)
            recency = 0.0 if total <= 1 else index / (total - 1)
            if index >= forced_recent_start:
                recency = max(recency, 0.85)
            score = relevance + self.history_recency_weight * recency
    # ... lines 169-301 omitted ...
    def compose(
        self,
        policy: list[dict[str, Any]],
        history: list[dict[str, Any]],
        evidence: list[Evidence],
        skills: list[Skill],
        tools: list[ToolSpec],
        budget: Budget,
        task: dict[str, Any],
    ) -> list[dict[str, Any]]:
        query_terms = _tokenize(str((task or {}).get("instruction") or ""))
        messages = [
            {"role": str(message.get("role") or "user"), "content": _content(message)}
            for message in policy
        ]
        policy_chars = sum(len(message["content"]) + _MESSAGE_OVERHEAD for message in messages)
        limit = budget.max_chars if budget.max_chars is not None else None
        if limit is None or policy_chars >= limit:
            return messages + [
                {"role": str(message.get("role") or "user"), "content": _content(message)}
                for message in history
            ]

        segments = self._memory_segments(
            evidence=evidence,
            skills=skills,
            query_terms=query_terms,
            tools=tools,
            start_order=-10,
        ) + self._history_segments(history=history, query_terms=query_terms)
        allocation = self._allocate(segments, max(0, limit - policy_chars))
        for segment in sorted(segments, key=lambda item: item.order):
            allowed = allocation.get(segment.order, 0)
            if allowed <= 0:
                continue
            content = _compress(segment.content, allowed, query_terms)
            if not content:
                continue
            messages.append({"role": segment.role, "content": content})
        return messages
# ... lines 342-346 omitted ...
\end{harnesscode}

\begin{harnesscode}{Skills: Skill-as-Pseudocode {\mdseries\itshape (Skill Compilation)}}{skil-p4.py}
# ... lines 1-48 omitted ...
def _skill_body(calls: tuple[str, ...], triggers: list[str], support: int) -> str:
    typed_steps = "\n".join(
        f"{index + 1}. emit Python that calls `apis.{call}(...)`"
        for index, call in enumerate(calls)
    )
    examples = "\n".join(
        f"- `apis.{call}(...)`"
        for call in calls
    )
    trigger_text = ", ".join(triggers) if triggers else "any overlapping task wording"
    return (
        "typed contract:\n"
        "- input.task_instruction: str\n"
        "- input.current_observation: str\n"
        "- output.python_action: str\n"
        f"- applies_when: task wording overlaps with {trigger_text}\n"
        f"- confidence_signal: seen in {support} successful training episodes\n"
        "algorithm:\n"
        f"{typed_steps}\n"
        "action template:\n"
        f"{examples}"
    )
        # ... lines 71-116 omitted ...
        for episode in solved:
            sequence = _sequence_from_trajectory(getattr(episode, "trajectory", ()))
            if not sequence:
                continue
            instruction = ""
            metadata = getattr(episode, "metadata", {}) or {}
            if isinstance(metadata, dict):
                instruction = str(metadata.get("instruction") or "")
            tokens = _tokenize(instruction)
            seen_here: set[tuple[str, ...]] = set()
            for width in range(1, min(self.max_ngram, len(sequence)) + 1):
                for start in range(0, len(sequence) - width + 1):
                    gram = tuple(sequence[start:start + width])
                    if gram in seen_here:
                        continue
                    seen_here.add(gram)
                    pattern_support[gram] += 1
                    bucket = pattern_triggers.setdefault(gram, Counter())
                    bucket.update(tokens)
        # ... lines 136-182 omitted ...
        for skill in self.skills:
            triggers = [str(item) for item in skill.get("triggers") or ()]
            overlap = _overlap_score(task_tokens, triggers)
            support = float(skill.get("support") or 0.0)
            score = overlap + min(0.25, support / 100.0)
            if score > best_score:
                best_skill = skill
                best_score = score
# ... lines 191-217 omitted ...
\end{harnesscode}

\begin{harnesscode}{Memories: Oracle Agent Memory {\mdseries\itshape (Structured Memory Architecture)}}{memo-p1.py}
"""Memories slot: task-specific experience retrieved by similarity.

Implements a lightweight version of the paper's memory lifecycle:
- ingest solved episodes via ``write`` / ``fit``
- extract durable summaries from trajectories
- consolidate them into a frozen JSON store via ``save``
- retrieve task-scoped memories by lexical similarity in ``recall``
"""
    # ... lines 9-75 omitted ...
    def _summarize_episode(self, episode_result: Any) -> str:
        instruction = str((episode_result.metadata or {}).get("instruction") or "").strip()
        lines: list[str] = []
        if instruction:
            lines.append(f"Task: {instruction}")
        for step in list(episode_result.trajectory or ())[:self.max_summary_steps]:
            code = str(step.get("code") or "").strip()
            output = str(step.get("output") or "").strip()
            if not code and not output:
                continue
            api_calls = [f"apis.{app}.{api}()" for app, api in _CALL_RE.findall(code)]
            action = ", ".join(api_calls) if api_calls else code.splitlines()[0][:120]
            detail = f"Step {step.get('step')}: {action}"
            if output:
                detail += f" -> {output.splitlines()[0][:120]}"
            lines.append(detail)
        if getattr(episode_result, "completed", False):
            lines.append("Outcome: completed successfully.")
        elif float(getattr(episode_result, "reward", 0.0) or 0.0) > 0.0:
            lines.append("Outcome: partially successful.")
        else:
            lines.append("Outcome: unsuccessful.")
        return "\n".join(lines)[:self.max_chars]
    # ... lines 99-111 omitted ...
    def recall(self, task: Any, budget: Any | None = None) -> list[Evidence]:
        instruction = str((task or {}).get("instruction") or "")
        wanted = _tokenize(instruction)
        if not wanted or not self._index:
            return []
        scored: list[tuple[float, dict[str, Any]]] = []
        for tokens, entry in self._index:
            similarity = _jaccard(wanted, tokens)
            if similarity < self.min_similarity:
                continue
            reward = float(entry.get("reward", 0.0) or 0.0)
            completed_bonus = 0.05 if entry.get("completed") else 0.0
            score = similarity + min(0.15, max(0.0, reward) * 0.05) + completed_bonus
            scored.append((score, entry))
        scored.sort(key=lambda item: (-item[0], str(item[1].get("task_id") or "")))
# ... lines 127-190 omitted ...
\end{harnesscode}

\begin{harnesscode}{Agentic workflows: TMAS {\mdseries\itshape (Multi-Agent Deliberation)}}{agen-p4.py}
"""TMAS-inspired agentic workflow slot.

This module keeps the slot boundary intact: it only controls intra-step reasoning flow.
It runs several specialist rollouts, aggregates them through a verifier, and carries
forward a tiny hierarchical memory built from prior step outputs:

* experience bank: concrete observations from earlier actions and outputs
* guideline bank: high-level strategy reminders to reduce redundant retries
"""
    # ... lines 10-129 omitted ...
    def decide(self, messages: list[dict[str, Any]], client: Any, state: Any = None) -> Proposal:
        self._remember(state)

        personas = (
            "Planner agent: propose a direct low-risk next action that gathers the missing fact or executes the task.",
            "Solver agent: propose the strongest task-completing next action using available APIs.",
            "Critic agent: propose a next action that avoids redundant work and fixes likely failure modes.",
        )
        memory = self._memory_text()
        candidate_texts: list[str] = []
        usage_total: dict[str, Any] = {}

        for persona in personas[:self.num_agents]:
            prompt = (
                persona
                + "\nReturn exactly one executable Python code block for the next environment action."
            )
            if memory:
                prompt += "\n\nUse this shared memory to coordinate with prior attempts:\n" + memory
            text, usage = self._complete([*messages, {"role": "user", "content": prompt}], client)
            candidate_texts.append(text)
            usage_total = _merge_usage(usage_total, usage)

        synthesis = [
            "Verifier agent: choose the best candidate action.",
            "Selection criteria: executable now, non-redundant, and most likely to advance the task.",
        ]
        if memory:
            synthesis.append("Shared memory:\n" + memory)
        for index, candidate in enumerate(candidate_texts, start=1):
            synthesis.append(f"Candidate {index}:\n{candidate}")
        synthesis.append("Return only the chosen next action as one Python code block.")

        final_text, usage = self._complete(
            [*messages, {"role": "user", "content": "\n\n".join(synthesis)}],
            client,
        )
        usage_total = _merge_usage(usage_total, usage)

        action = _extract_code(final_text)
        if not action:
            for candidate in candidate_texts:
                action = _extract_code(candidate)
                if action:
                    final_text = candidate
                    break

# ... lines 177-196 omitted ...
\end{harnesscode}

\end{document}